\documentclass{article}

 \usepackage[final]{neurips_2026}
\makeatletter
\providecommand{\@trackname}{}
\makeatother

\makeatletter
\providecommand{\@trackname}{}
\makeatother

\usepackage[utf8]{inputenc} 
\usepackage[T1]{fontenc}    
\usepackage{hyperref}       
\usepackage{url}            
\usepackage{booktabs}       
\usepackage{amsfonts}       
\usepackage{nicefrac}       
\usepackage{microtype}      
\usepackage{xcolor}         

\usepackage{siunitx}
\usepackage{amsmath}
\usepackage[newfloat,frozencache,cachedir=minted-cache]{minted}
\usepackage{caption}
\usepackage{graphicx}
\usepackage{indentfirst}
\usepackage{makecell}
\usepackage{xcolor}
\usepackage{subcaption}
\usepackage{longtable}
\usepackage{multirow}
\usepackage{tabularx}

\setcitestyle{numbers,square}

\graphicspath{{figures/}}

\SetupFloatingEnvironment{listing}{name=Listing,within=section}

\definecolor{lstbg}{gray}{0.97}

\setminted{
  fontsize=\footnotesize,
  breaklines=true,
  numbersep=6pt,
  tabsize=2,
  frame=lines,
  framesep=5pt,
  bgcolor=lstbg,
  escapeinside=@@,
}

\newminted[cudacode]{cuda}{}
\newminted[metalcode]{cpp}{}
\newmintinline[ccode]{cuda}{}
\newmintinline[mcode]{cpp}{}

\newcommand{\Transfers}{\textsc{transfers}}
\newcommand{\HwGap}{\textsc{hw-gap}}
\newcommand{\RuntimeGone}{\textsc{runtime}}
\newcommand{\NotApplicable}{\textsc{n/a}}

\newcommand{\std}[1]{{\scriptsize\color{gray}$\pm$ #1}} 

\newcommand{\pct}[1]{{\scriptsize\color{gray}(#1)}} 

\DeclareSIUnit{\tok}{tok}
\newcommand{\gbs}[1]{\SI{#1}{\giga\byte\per\second}}

\newcommand{\us}[1]{\SI{#1}{\micro\second}}
\newcommand{\x}[1]{$#1\times$}

\newcommand{\speedup}{S}

\title{Evaluating the Applicability of LLM Inference Optimizations on Apple Silicon} 

\author{%
  Kyrylo Yemets\textsuperscript{1,2} \quad
  Andrii Yaroshevych\textsuperscript{1} \quad
  Victor Muryn\textsuperscript{1} \quad
  Ostap Khomenko\textsuperscript{1} \\
  \textbf{Pavlo Kryven\textsuperscript{1} \quad
  Maksym Kmet\textsuperscript{1} \quad
  Maksym Shamrai\textsuperscript{1, 3} \quad
  Mariya Hirna\textsuperscript{1}} \\[4pt]
  \textsuperscript{1}MacPaw AI\&Research, Kyiv, Ukraine \\
  \textsuperscript{2}Department of Artificial Intelligence,
  Lviv Polytechnic National University, Lviv, Ukraine \\
  \textsuperscript{3} Institute of Mathematics of NAS of Ukraine, Kyiv, Ukraine \\[4pt]
  \texttt{\{ruiite, andrii.yaroshevych, victormuryn, ostap.khomenko,} \\
  \texttt{pavlo.kryven, maxkmet, mshamrai, maryhirna\}@macpaw.com}
}

\begin{document}

\maketitle

\begin{abstract}
Apple Silicon is among the most widely used platforms for local LLM inference, yet nearly all inference optimizations are designed and validated on server-class NVIDIA GPUs under batched serving workloads. Which optimizations retain their benefits when moved to Apple Silicon, and what determines their applicability? We evaluate four optimization families: KV-cache pruning, speculative decoding, ported GPU kernels, and post-training quantization. Experiments span four Apple machines under a common single-request decode workload, with matched RTX~4090 comparisons where possible. KV-cache pruning can reach up to $1.75\times$ end-to-end throughput on Apple but can reduce throughput on desktop NVIDIA GPUs. Under the common benchmark, all evaluated speculative decoding methods reduce throughput on Apple, despite reported $2$--$6\times$ gains on server GPUs. CUDA kernel designs transfer only partially: public Metal lacks asynchronous copy, and batch-one decode makes several serving-oriented mechanisms inapplicable, although transferable components yield up to $1.20\times$ decode throughput. Quantization gains vary by machine. Across four families, a reported speedup is not a portable property of a technique alone but a joint function of technique, runtime, hardware contract, and workload regime.
\end{abstract}

\section{Introduction}

Local LLM inference increasingly happens on Apple machines: llama.cpp treats M-series chips as a first-class target \cite{llamacpp}, Apple ships MLX for unified-memory inference \cite{hannun_mlx_2023}, and recent studies describe M-series unified memory as the practical consumer path to large open-weight models \cite{javat2026silicon}. The optimizations that would make this inference fast, however, carry no evidence for it: none of the methods we study --- KV-cache management \cite{kwon_efficient_2023}, speculative decoding \cite{leviathan2023fast}, the FlashAttention/FlashInfer, Marlin, and MegaBlocks kernels \cite{dao_flashattention_2022,ye_flashinfer_2025,frantar_marlin_2024,gale_megablocks_2023}, and post-training quantization \cite{frantar_gptq_2023} --- reports a measurement on Apple hardware in its original publication. 

An optimization that reduces memory traffic or computation may still fail to improve end-to-end throughput if it targets a secondary cost, adds runtime overhead, or relies on hardware features or workload conditions that are absent from local inference. This motivates our central question: \emph{Which LLM inference optimizations remain effective on Apple Silicon, and why?} 

Our experiments focus on the single-request, unified-memory regime that exists at consumer scale, and we evaluate the bundle users actually deploy: hardware, runtime, and workload together. We cover four Apple devices across three GPU generations, dense and hybrid architectures, and matched RTX~4090 controls where configurations can be aligned.

Prior Apple-Silicon work has mainly evaluated models or complete software stacks. MLX provides a framework for unified-memory inference \cite{hannun_mlx_2023}, OpenELM reports local generation with MLX \cite{mehta_openelm_2024}, and \citet{rajesh_production-grade_2025} compare several runtimes on an M2~Ultra. That study establishes performance trade-offs among Apple-Silicon devices. In contrast, we isolate individual optimizations, port representative CUDA mechanisms to Metal/MLX, and measure their kernel-level and end-to-end effects.

\textbf{Contributions.} 
\textbf{(1)} A controlled comparison of four optimization families for local inference on Apple Silicon. 
\textbf{(2)} Metal/MLX ports of representative CUDA attention, quantized-matmul, and MoE mechanisms, evaluated at both kernel and end-to-end levels. 
\textbf{(3)} A cross-platform analysis explaining when gains persist, disappear, or reverse because of hardware, runtime, and workload differences.








\section{Methods and Results}



Table~\ref{tab:methods-overview} summarizes the four optimization families evaluated in this paper: the techniques tested within each, the Apple Silicon generations used, and the key finding.
The subsections that follow detail the experimental setup, results, and causal analysis for each area.

\begin{table}[ht]
\centering
\caption{Overview of the four optimization families evaluated in this paper. NVIDIA RTX~4090 is used as a matched control where noted.}
\label{tab:methods-overview}
\small
\setlength{\tabcolsep}{2pt}
\renewcommand{\arraystretch}{0.90}
\begin{tabularx}{\textwidth}{>{\raggedright\arraybackslash}p{0.14\textwidth}
                              >{\raggedright\arraybackslash}X
                              >{\raggedright\arraybackslash}p{0.15\textwidth}
                              >{\raggedright\arraybackslash}X}
\toprule
\textbf{Optimization} & \textbf{Methods} & \textbf{Devices} & \textbf{Key Finding} \\
\midrule
KV Cache
  & Prefix reuse~\cite{mlx2026mlx-lm}; KVPress~\cite{nvidia2026kvpress} pruning (ExpectedAttention~\cite{devoto2025expectedattention}, FastKVzip~\cite{kim2026fastkvzip}, and 4 others)
  & M2 Pro, M3 Ultra, M4 Pro; RTX~4090
  & \textbf{Prefix reuse:} $-57\%$ wall time at turn~9.
    \textbf{Pruning:} speedup of up to $1.75\times$ on Apple Silicon vs up to $1.02\times$ on RTX~4090. \\
\addlinespace
Speculative Decoding
  & Classic~\cite{leviathan2023fast}, N-gram, Cascade-n-gram~\cite{chen2024cascade}, EAGLE-3~\cite{li2025eagle3}, DFlash~\cite{chen2026dflash}, DDTree~\cite{ringel2026ddtree}
  & M2 Pro, M3 Ultra
  & \textbf{No method achieved a net speedup.}\newline
    Reported $2$--$6\times$ gains on NVIDIA GPUs do not generally transfer to different Apple GPUs.\\
\addlinespace
Metal Kernels
  & Marlin~\cite{frantar_marlin_2024} (W4A16 GEMM), FlashInfer~\cite{ye_flashinfer_2025} (split-KV attention), MoE Gather~\cite{gale_megablocks_2023}
  & M2, M2 Pro, M3 Ultra, M4 Pro; RTX~4090
  & \textbf{MoE Gather}: $+3$--$12\%$ universally.\newline
    \textbf{FlashInfer}: $+20\%$ on M4 Pro, $-3\%$ on M2.
    \textbf{Marlin}: effective at batch ${>}1$, negligible at $M{=}1$. \\
\addlinespace
Quantization
  & GGUF/llama.cpp; MLX~\cite{hannun_mlx_2023} (affine, DWQ); RTN, QuaRot+RTN~\cite{ashkboos_quarot_2024}, QuaRot+GPTQ~\cite{frantar_gptq_2023} 
  & M2 Pro, M3 Ultra; RTX~4090
  & \textbf{MLX native affine/DWQ} is the best combined trade-off.
  \textbf{QuaRot} does not pay off on Apple Silicon without low-precision activation kernels.\\
\bottomrule
\end{tabularx}
\vspace{-0.8\baselineskip}
\end{table}


\subsection{KV Cache}

\label{sec:kv-cache}

KV caching reduces autoregressive decoding cost by retaining the keys and values computed for preceding tokens. This avoids repeated computation but introduces memory-footprint and bandwidth costs that increase with context length~\citep{kwon2023pagedattention}. We characterize cache growth and evaluate the latency and memory effects of persistent prefix reuse and post-prefill pruning. Evaluating pruning-induced changes in output quality is outside our scope; these effects are expected to be comparable across platforms using identical pruning algorithms and configurations.

\subsubsection{Experimental setup}
\label{sec:kv-setup}

Cache footprints are derived from the released configurations of \texttt{mlx-community/Qwen3-4B-4bit}, \texttt{mlx-community/gemma-3-4b-it-4bit}, \texttt{mlx-community/granite-4.0-h-micro-4bit}, and \texttt{LiquidAI/LFM2.5-2.6B}, together with the MLX-LM cache layout~\citep{mlx2026mlx-lm}. Cached tensors are assumed to use bfloat16, independently of checkpoint weight precision. Pruning experiments use Qwen3-4B-Instruct-2507 and LFM2.5-2.6B in bfloat16 through Transformers and KVPress~\citep{wolf2020transformers,nvidia2026kvpress}. Qwen is evaluated on M2~Pro, M3~Ultra, M4~Pro, and NVIDIA RTX~4090; LFM is evaluated on the three Apple devices. The matched workloads use context lengths of 512 and 4,096 tokens and output lengths of 1, 16, 128, and 256 tokens at batch size one.

Timings are medians of three measured runs after one warm-up run. We define speedup as $\speedup = T_{\mathrm{unpruned}}/T_{\mathrm{pruned}}$, using the baseline for the same device and workload; values above one indicate  lower end-to-end latency. These ratios only compare the effectiveness of pruning and are not indicative of absolute throughput. Peak memory is reported on M4~Pro, whose latency trends are broadly representative of the Apple measurements. Positive memory savings indicate a lower process-level peak than the unpruned baseline.

The experiments include SnapKV, KNorm, StreamingLLM, and TOVA methods at a compression ratio of 0.5 and ExpectedAttention at 0.75~\citep{li2024snapkv,devoto2024knorm,xiao2024streamingllm,oren2024tova,devoto2025expectedattention}. The ratio denotes the fraction of cached entries removed. FastKVzip at 0.75 is included for Qwen but requires pretrained model-specific gates unavailable for LFM2.5-2.6B~\citep{kim2026fastkvzip}. LFM2.5-8B-A1B is excluded because the KVPress integration does not support its \texttt{Lfm2MoeForCausalLM} architecture; its bfloat16 weights also exceed the available memory on M2~Pro, precluding a matched three-device comparison~\citep{liquidAI20268BA1B}.

\subsubsection{Cache growth and persistent prefix reuse}
\label{sec:kv-footprint-reuse}

The persistent cache footprint depends on attention architecture and retention policy rather than parameter count alone. At batch size one, it can be expressed as
\begin{equation}
    M_{\mathrm{cache}}(n)
    = 2b \sum_{\ell\in\mathcal{A}}
      h^{\mathrm{KV}}_{\ell} d_{\ell}\min(n,w_{\ell})
    + M_{\mathrm{fixed}},
    \label{eq:kv-footprint}
\end{equation}
where $n$ is context length, $b$ is the cache scalar width in bytes, $\mathcal{A}$ is the set of attention layers, and $h^{\mathrm{KV}}_{\ell}$ and $d_{\ell}$ are the number and dimensionality of KV heads in layer $\ell$. The retention window $w_{\ell}$ is finite for sliding-window attention and infinite for global attention. The factor of two accounts for separate key and value tensors, while $M_{\mathrm{fixed}}$ includes sequence-length-independent recurrent or convolutional state.

\begin{figure}[!htbp]
    \centering
    \includegraphics[width=\textwidth]{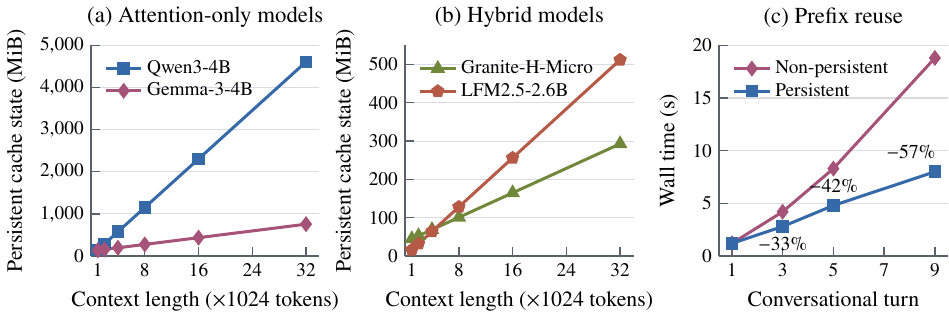}
    \caption{Cache growth and persistent prefix reuse. Panels (a) and (b) report architecture-derived persistent cache state under the bfloat16 MLX-LM layout. Gemma-3-4B uses 1,024-token local-attention caches; hybrid-model values include fixed recurrent or convolutional state. Panel (c) reports measured Qwen3-4B wall time with non-persistent and persistent conversational caches.}
    \label{fig:kv-cache-scaling}
\end{figure}

Figure~\ref{fig:kv-cache-scaling}(a) applies Equation~\ref{eq:kv-footprint} to the attention-only models. Qwen3-4B contains 36 global-attention blocks with eight 128-dimensional KV heads, giving 144~MiB at 1,024 tokens and 4,608~MiB at 32,768 tokens~\citep{qwen2025qwen3}. Gemma-3-4B contains 34 attention blocks with four 256-dimensional KV heads, but only five use global attention; the other 29 retain at most 1,024 tokens under the evaluated MLX cache policy. Its footprint therefore increases from 136 to 756~MiB over the same range~\citep{gemmateam2025gemma3,mlx2026mlx-lm}.

The hybrid models retain less token-indexed attention state (Figure~\ref{fig:kv-cache-scaling}(b)). Granite-4.0-H-Micro combines four global-attention blocks with 36 Mamba2 blocks, whereas LFM2.5-2.6B combines eight global-attention blocks with 22 short-convolution blocks~\citep{ibm2025granite4,liquid2025lfm2}. Their token-indexed state grows by 8 and 16~KiB per token, respectively; fixed recurrent and convolutional states contribute 36.9 and 0.17~MiB. At 32,768 tokens, their footprints are consequently 292.9 and 512.2~MiB. Fewer attention blocks reduce cache growth and the state eligible for pruning.

Persistent caching extends reuse beyond a single response by retaining the longest unchanged prefix across requests, avoiding repeated prefill~\citep{mlx2026mlx-lm,zheng2024sglang}. In the MLX-LM conversation, wall time at turns 3, 5, and 9 falls by 33\%, 42\%, and 57\%, respectively (Figure~\ref{fig:kv-cache-scaling}(c)). These savings reflect avoided prefill rather than increased steady-state decoding throughput.

\subsubsection{Dense-model pruning across devices}
\label{sec:kv-qwen}

\begin{figure}[!htbp]
    \centering
    \includegraphics[width=\textwidth]{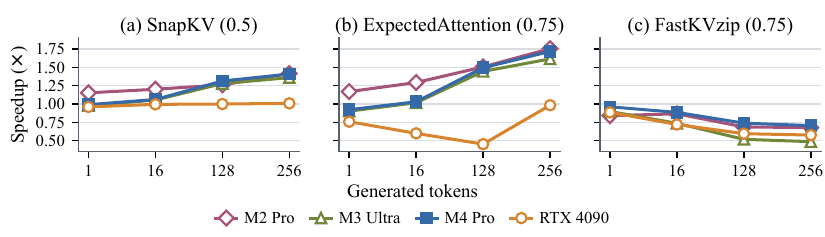}
    \caption{Device-normalized KV-pruning speedup for Qwen3-4B-Instruct-2507 in bfloat16 with a 4,096-token context. Each curve uses the unpruned baseline on the same device. Values in parentheses denote compression ratios.}
    \label{fig:kv-hardware-comparison}
\end{figure}

Figure~\ref{fig:kv-hardware-comparison} compares three representative pruning methods at a 4,096-token context. SnapKV represents KNorm, StreamingLLM, and TOVA, which differ in retention criteria but share post-prefill token selection and cache compaction and exhibit similar measured trends. ExpectedAttention shows the strongest hardware dependence, whereas FastKVzip provides an overhead-dominated counterexample. At 128--256 generated tokens, SnapKV and ExpectedAttention achieve 1.26--1.75$\times$ speedup on Apple Silicon, versus 0.45--1.01$\times$ on the RTX~4090; FastKVzip remains below baseline on all devices (0.49--0.74$\times$). At 1--16 generated tokens, M2~Pro attains the largest speedups for SnapKV and ExpectedAttention, consistent with a larger relative contribution of the eliminated attention work. This separation diminishes with output length, and the ratios are not ordered by chip generation.

\begin{figure}[!htbp]
    \centering
    \includegraphics[width=\textwidth]{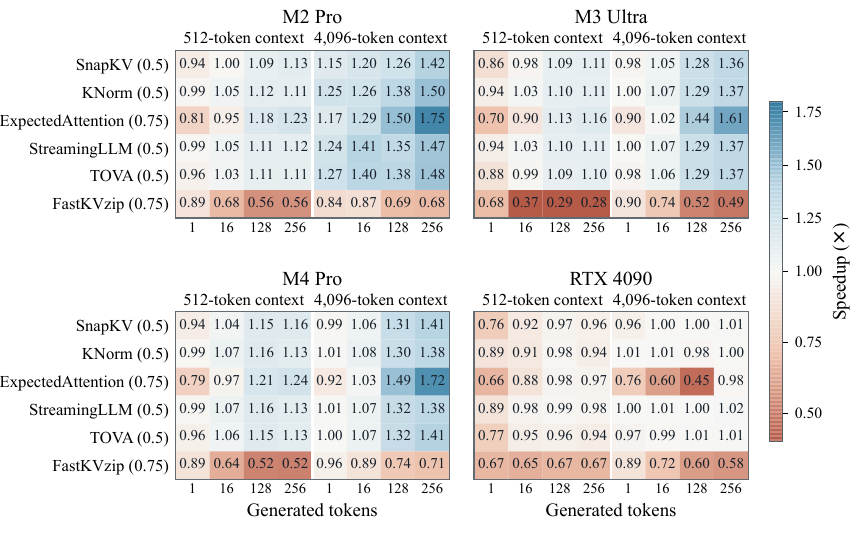}
    \caption{Complete matched end-to-end speedup grid for Qwen3-4B-Instruct-2507 in bfloat16. Columns vary context and output length; each cell uses the unpruned baseline for the same device and workload. Values above one indicate a latency reduction.}
    \label{fig:kv-qwen-speedup}
\end{figure}

The complete matched grid shows how this hardware dependence varies with context length (Figure~\ref{fig:kv-qwen-speedup}). At a 4,096-token context and 256-token output, ExpectedAttention attains 1.75$\times$, 1.61$\times$, and 1.72$\times$ on M2~Pro, M3~Ultra, and M4~Pro, respectively, but 0.98$\times$ on the RTX~4090. SnapKV, KNorm, StreamingLLM, and TOVA yield 1.36--1.50$\times$ across the Apple devices and 0.99--1.02$\times$ on the RTX~4090 in this workload. At a 512-token context, the maximum Apple speedup is only 1.24$\times$. FastKVzip remains below baseline throughout the grid, at 0.28--0.96$\times$. These results are consistent with reduced attention work more effectively amortizing pruning overhead in longer Apple workloads; compression ratio alone does not determine the resulting speedup.

\subsubsection{Hybrid-model pruning and peak memory}
\label{sec:kv-lfm}

\begin{figure}[!htbp]
    \centering
    \includegraphics[width=\textwidth]{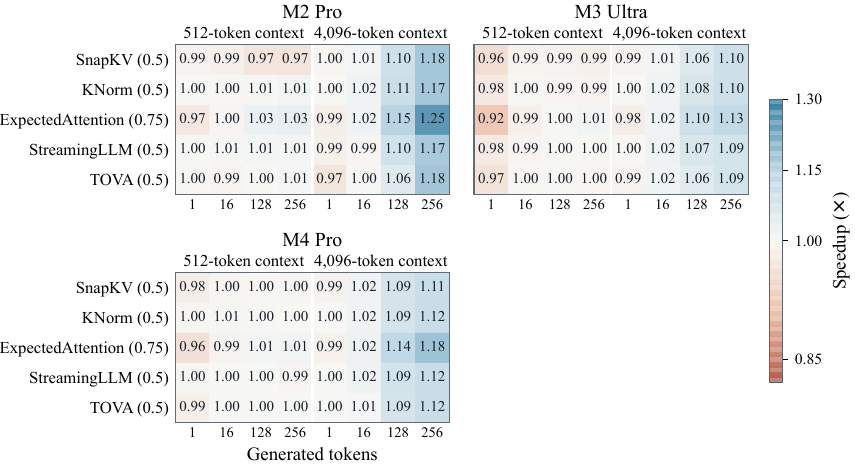}
    \caption{Device-normalized end-to-end speedup for LFM2.5-2.6B in bfloat16. The matched grid contains five pruning methods evaluated on all three Apple devices.}
    \label{fig:kv-lfm-speedup}
\end{figure}

The smaller amount of prunable attention state in LFM2.5-2.6B is accompanied by smaller latency gains (Figure~\ref{fig:kv-lfm-speedup}). At a 512-token context, every method remains between 0.92$\times$ and 1.03$\times$ across the three Apple devices. At a 4,096-token context and 256-token output, speedups range from 1.17--1.25$\times$ on M2~Pro, 1.09--1.13$\times$ on M3~Ultra, and 1.11--1.18$\times$ on M4~Pro, with ExpectedAttention highest on each device. The maximum is thus 1.25$\times$, compared with 1.75$\times$ for Qwen3-4B-Instruct-2507. The output-length dependence is qualitatively similar to Qwen, but attention-only pruning leaves LFM's 22 short-convolution blocks and their state unchanged, limiting the inference cost it can remove.

\begin{figure}[!htbp]
    \centering
    \includegraphics[width=\textwidth]{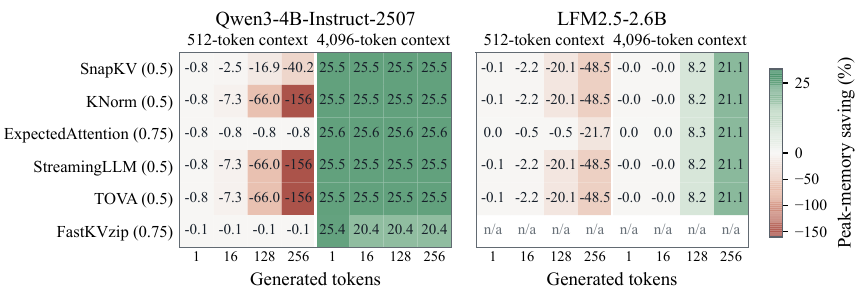}
    \caption{Process-level peak-memory saving on M4~Pro relative to the corresponding unpruned workload. Positive values indicate a lower peak; negative values indicate an increase. FastKVzip is unavailable for LFM2.5-2.6B.}
    \label{fig:kv-memory}
\end{figure}

\label{sec:kv-memory}
Process-level peak memory also depends on temporary allocations rather than retained-cache size alone (Figure~\ref{fig:kv-memory}). On M4~Pro at a 4,096-token context, Qwen peak-memory savings range from 20.4\% to 25.6\% across the reported methods and outputs. LFM2.5-2.6B shows no material reduction for outputs of 1 or 16 tokens, followed by savings of 8.2--8.3\% at 128 tokens and 21.1\% at 256 tokens. At a 512-token context, pruning instead increases peak memory by up to 155.5\% for Qwen and 48.5\% for LFM. These increases are consistent with transient scoring and compaction allocations outweighing the smaller persistent cache. Retained KV state, process-level peak memory, and end-to-end latency therefore measure distinct effects of pruning.

Persistent prefix reuse yields consistent latency reductions by eliminating redundant prefill without token-selection overhead. Pruning gains depend on context and output length, device, and the fraction of inference cost attributable to attention KV state. In the matched Qwen workloads, the largest gains occur on Apple Silicon, while the RTX~4090 remains close to baseline or slows down; FastKVzip slows execution on both platforms. Neither compression ratio nor processor generation alone predicts end-to-end benefit, and reduced KV state does not ensure lower process-level peak memory.


\subsection{Speculative Decoding}

\label{sec:speculative-decoding}
\label{app:speculative-prior}

Speculative decoding accelerates autoregressive generation by separating token proposal from verification~\cite{leviathan2023fast}. A cheaper drafting procedure proposes a sequence of candidate tokens, which the target model evaluates in a single forward pass. Verification retains the accepted prefix and produces the next target token, allowing several output tokens to be generated per target-model invocation. Research on speculative decoding has developed this principle by reducing drafting cost, improving the agreement between draft and target predictions, and increasing the parallelism of token proposal.

The classical approach, represented by Classic in our evaluation, uses a separate, smaller autoregressive language model to generate the draft sequence. The target then verifies these proposals together, without requiring retraining. The original method reports \(2\)--\(3\times\) acceleration for T5-XXL relative to standard T5X inference~\cite{leviathan2023fast}; a later greedy-decoding evaluation reports \(1.93\times\) on MT-Bench with Vicuna-13B~\cite{li2025eagle3}. Although the auxiliary model is smaller, its sequential drafting steps still contribute to generation latency.

One way to reduce this cost is to replace neural drafting with inexpensive statistical lookup. N-gram drafting retrieves stored continuations of a matching token suffix, avoiding a neural forward pass. Such a procedure can also assist a neural drafter: Cascade Speculative Drafting recursively accelerates each drafter with a cheaper one, terminating at the statistical Max-Gram model~\cite{chen2024cascade}. Our Cascade-n-gram evaluates this vertical cascade using N-gram proposals for a small neural drafter. The full published method additionally assigns cheaper drafters to later positions and achieves \(1.70\times\) speedup for Vicuna-7B on GSM8K (\(56.72\) versus \(33.34\) tokens/s)~\cite{chen2024cascade}.

A complementary direction improves draft quality by using features already computed by the target model. The EAGLE family introduces lightweight learned drafting modules conditioned on these features. EAGLE-3 predicts tokens from fused low-, middle-, and high-layer representations, with a training-time test that feeds back the drafter's own outputs to improve successive predictions; inference can use a dynamic candidate tree. Under greedy decoding, it reports \(5.58\times\) and \(4.40\times\) on MT-Bench with Vicuna-13B and LLaMA-3.1-8B-Instruct, respectively, and \(6.47\times\) on HumanEval with Vicuna-13B~\cite{li2025eagle3}.

Feature conditioning improves proposal quality, but autoregressive drafting still requires sequential forward passes. More recent work therefore explores generating draft tokens in parallel. DFlash combines target-feature conditioning with a block diffusion drafter that predicts masked positions together in one pass, reporting \(2.85\times\) on MT-Bench and \(6.09\times\) on MATH-500 with Qwen3-4B~\cite{chen2026dflash}. DDTree extends this approach by constructing a candidate tree from the same per-position distributions under a node budget, maximizing a surrogate for expected acceptance length before joint verification. In its separate evaluation, DDTree improves DFlash's MT-Bench speedup from \(2.64\times\) to \(4.18\times\), and its MATH-500 speedup from \(5.54\times\) to \(7.50\times\)~\cite{ringel2026ddtree}. These published gains motivate our evaluation, although their models, workloads, and runtimes differ from ours; the learned and block-drafting studies primarily use server-class NVIDIA GPUs and 16-bit targets.

\subsubsection{Matched Apple Silicon results}
\label{sec:speculative-matched}

We evaluate Classic, N-gram, Cascade-n-gram, EAGLE-3, DFlash, and DDTree to test whether the benefits of these drafting strategies persist in the single-request MLX regime motivated in the introduction. The principal comparison uses M2 Pro and M3 Ultra devices with a 4-bit Qwen3-4B target, representative of local inference, and a matched BF16 target to examine precision sensitivity. All six methods use the same 100 conversational prompts, batch size and concurrency of one, and greedy generation, with identical method-specific settings across the two devices. Prompts average \(1{,}776\) tokens, whereas baseline completions average approximately \(54\)--\(62\) tokens. This distribution is chosen to represent local chat and tool use with small models, where system prompts and conversational context produce relatively long inputs followed by short responses. Each MLX adaptation is normalized to its own autoregressive baseline; absolute throughput across implementations is therefore not a controlled cross-method comparison. Appendix~\ref{app:speculative-implementation} gives the checkpoints, draft budgets, and measurement protocol.

\begin{figure}[htbp]
    \centering
    \includegraphics[width=\linewidth]{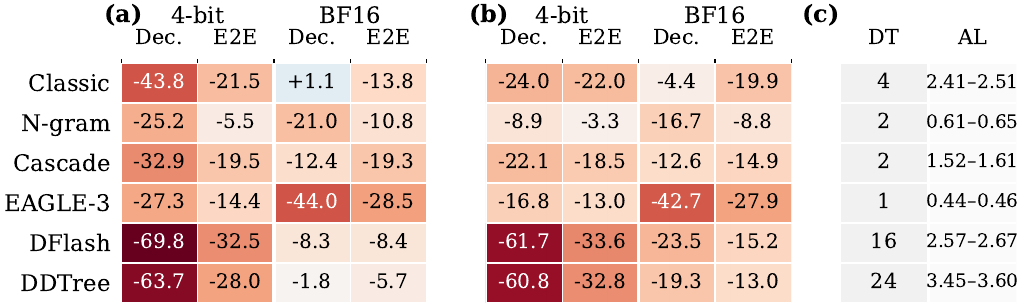}
    \caption{Decode (Dec.) and standardized end-to-end (E2E) throughput changes (\%) relative to each implementation's autoregressive baseline for Qwen3-4B on (a) M2 Pro and (b) M3 Ultra. Panel (c) gives the configured draft budget (DT) and mean accepted draft length (AL), shown as a range across the two devices for the 4-bit target. For DFlash, DT is the 16-position block including the target-produced root; for DDTree, DT is the budget of 24 non-root candidate nodes. AL excludes the root for both methods.}
    \label{fig:spec_decoding}
\end{figure}

No configuration in this matched comparison improves E2E throughput (Figure~\ref{fig:spec_decoding}). With the 4-bit target, all six methods also reduce decode throughput: losses range from \(25.2\%\) to \(69.8\%\) on M2 Pro and from \(8.9\%\) to \(61.7\%\) on M3 Ultra. N-gram is closest to E2E break-even, at \(-5.5\%\) and \(-3.3\%\), respectively, despite its approximately near-zero lookup cost. A slower target baseline alone does not make speculation beneficial: even N-gram's decode penalty is larger on M2 Pro, indicating that the complete draft--verify path scales differently from an autoregressive step.

BF16 likewise yields no E2E gain. Classic on M2 Pro is the only decode-positive configuration (\(+1.1\%\)), but its E2E throughput falls by \(13.8\%\). DFlash and DDTree show the strongest precision sensitivity: their decode penalties fall from \(69.8\%\) and \(63.7\%\) to \(8.3\%\) and \(1.8\%\) on M2 Pro, and from \(61.7\%\) and \(60.8\%\) to \(23.5\%\) and \(19.3\%\) on M3 Ultra. On M2 Pro, this mainly reflects a slower baseline: moving to BF16 reduces autoregressive decode throughput from approximately \(56.7\) to \(21.0\) tokens/s, while DFlash and DDTree remain near \(19\)--\(21\) tokens/s. EAGLE-3 instead becomes slower relative to its baseline on both devices, so target precision does not affect all methods uniformly.

Taken together, these results show that neither inexpensive token proposals nor a more costly target step consistently produces a net gain. To examine why the methods remain slower, we relate the number of tokens accepted in each round to the total cost of drafting and verification.

\paragraph{Acceptance and decoding cost.}
In the 4-bit M3 Ultra setting, DFlash accepts a mean of \(2.57\) draft tokens per round and DDTree \(3.45\), excluding the root (equivalently, \(3.57\) and \(4.45\) with the root). Nevertheless, DFlash reduces decode throughput from \(156.7\) to \(60.0\) tokens/s, and DDTree reaches only \(61.4\) tokens/s. EAGLE-3 and Cascade-n-gram similarly fall from \(261.2\) to \(217.3\) and from \(162.0\) to \(126.3\) tokens/s, respectively. Classic accepts \(2.41\)--\(2.51\) draft tokens per round across the two devices, yet on M2 Pro falls from \(56.65\) to \(31.86\) tokens/s. These within-implementation comparisons show that accepted work must be evaluated against the cost of the entire speculative round.

An approximate decode speedup is
\begin{equation}
    S_{\mathrm{dec}} \approx
    \frac{\tau\,t_{\mathrm{AR}}}
    {t_{\mathrm{draft}}+t_{\mathrm{tree}}+
     t_{\mathrm{verify}}(B)+t_{\mathrm{commit}}},
    \label{eq:speculative-cost}
\end{equation}
where \(\tau\) is the mean number of newly committed output tokens per round, including the target-produced correction or bonus token as applicable; \(t_{\mathrm{AR}}\) is the latency of one autoregressive target step; and the denominator includes drafting, tree construction, verification at width \(B\), and output/KV-cache commitment. For methods without a tree, \(t_{\mathrm{tree}}=0\). Thus \(\tau\) is distinct from the accepted-draft length AL in Figure~\ref{fig:spec_decoding}. Decoding accelerates only when the total round cost is smaller than \(\tau t_{\mathrm{AR}}\); increasing acceptance or the draft budget alone does not ensure this inequality. This trade-off motivates the following analysis of how workload predictability, target cost, and verification overhead affect decode performance, and how prefill limits the resulting E2E benefit.

\subsubsection{Factors limiting speedup}
\label{sec:speculative-limits}

\paragraph{Workload predictability.}
The published block-drafting results show substantially less amortizable work on chat than on mathematical reasoning. For Qwen3-4B, DFlash's reported acceptance length, including the target-produced token, falls from \(7.84\) on MATH-500 to \(4.35\) on MT-Bench, while speedup falls from \(6.09\times\) to \(2.85\times\)~\cite{chen2026dflash}. DDTree similarly falls from \(10.71\) to \(6.70\) tokens and from \(7.50\times\) to \(4.18\times\)~\cite{ringel2026ddtree}. Our conversational workload is consistent with this less predictable regime. These cross-workload observations support predictability as a limiting factor, without isolating it from differences in checkpoints, precision, and implementation.

\paragraph{Small and quantized targets.}
A 4-bit 4B target usual for the local on-device inference already has inexpensive autoregressive steps, reducing the work that a speculative round can replace. Draft-side costs need not shrink proportionally: DFlash and DDTree retain a BF16 drafter~\cite{zlab2026qwen3dflash}, and the EAGLE-3 head remains 4-bit even with the BF16 target. The BF16 comparison is therefore a sensitivity test of target precision within each method's pipeline. The supplementary 8B-4bit experiment also improves relative E2E performance for most method--device pairs, but none reaches break-even; N-gram comes closest at \(-1.4\%\) on M3 Ultra (Appendix~\ref{app:speculative-8b}). More expensive target steps improve the trade-off for several methods, but neither target size nor quantization alone explains the negative results.

\paragraph{Backend-dependent verification.}
The wider target pass is only part of a speculative round. Block or tree preparation, custom attention masks, verifier traversal, KV-cache updates, and synchronization can add costs that do not scale with single-token target throughput. DFlash and DDTree's smaller BF16 decode deficits on M2 Pro than on M3 Ultra illustrate this mismatch. A separate A100 DDTree reproduction achieves \(3.20\times\) decode speedup despite using PyTorch scaled dot-product attention (SDPA), rather than FlashAttention, for the ancestor-only mask. Consequently, the observed gap concerns the complete MLX and PyTorch draft--verify paths. Without kernel-level profiling or matched backend ablations, these results do not establish a hardware-only cause or a specific Metal-kernel limitation.

\paragraph{Non-accelerated prefill.}
The long prompts and short completions make prefill a substantial part of E2E latency. Speculation mainly accelerates decode and can add draft-model prefill, feature extraction, and cache initialization. With the BF16 target on M3 Ultra, Classic and Cascade-n-gram reduce prefill throughput by \(29.6\%\) and \(16.8\%\), respectively, compared with \(0.7\%\) for N-gram. If baseline prefill occupies a fraction \(f_{\mathrm{pre}}\) of total latency, unchanged prefill imposes the idealized ceiling
\begin{equation}
    S_{\mathrm{e2e}} \leq \frac{1}{f_{\mathrm{pre}}},
    \label{eq:speculative-prefill-bound}
\end{equation}
even with cost-free decode; additional prefill overhead lowers it. In a separate M3 Pro DDTree run, baseline prefill and decode take \(2.60\) and \(3.39\) seconds per request, giving a ceiling near \(2.3\times\), while speculative prefill rises to \(2.77\) seconds. A matched A100 experiment using the same dataset and target family reports mean time to first token of \(0.109\) seconds versus \(2.66\) seconds on M3 Pro, consistent with a larger prefill limitation in the tested Apple stack; the different runtimes preclude hardware-only attribution.

Shorter prompts and longer outputs would reduce prefill's share, but would not by themselves reverse the measured decode and E2E losses. In the principal comparison, only M2 Pro BF16 Classic has a positive decode change, and that gain is \(1.1\%\). Other length regimes were not directly evaluated and may also change acceptance and phase throughput. The negative result therefore applies to the measured configurations, while the cost model explains the conditions needed for a gain.

\subsubsection{Configuration-specific positive cases}
\label{sec:speculative-positive}

Supplementary experiments identify configurations with positive decode or E2E changes (Table~\ref{tab:speculative-other-apple}). The M2 Pro BF16 Classic row repeats the decode-only exception in the principal comparison; the remaining rows extend the device coverage. These observations use separate implementations and are not a controlled ranking of methods or processors.

\begin{table}[htbp]
    \centering
    \small
    \setlength{\tabcolsep}{3.5pt}
    \caption{Apple Silicon configurations with an observed positive change in decode or standardized E2E throughput relative to the paired autoregressive baseline.}
    \label{tab:speculative-other-apple}
    \begin{tabular}{@{}lllrr@{}}
        \toprule
        Method & Device & Target & E2E & Decode \\
        \midrule
        Classic       & M2 Pro & Qwen3-4B-BF16 & \(-13.8\%\) & \(+1.1\%\) \\
        Classic       & M4 Pro & Qwen3-4B-4bit & \(-5.0\%\)  & \(+10.0\%\) \\
        N-gram        & M3 Max & Qwen3-4B-4bit & \(+2.0\%\)  & \(+13.0\%\) \\
        Cascade-n-gram & M3 Max & Qwen3-4B-4bit & \(-11.6\%\) & \(+7.6\%\) \\
        DFlash        & M3 Pro & Qwen3-4B-BF16 & \(+9.2\%\)  & \(+25.0\%\) \\
        DDTree        & M3 Pro & Qwen3-4B-BF16 & \(+15.5\%\) & \(+40.4\%\) \\
        \bottomrule
    \end{tabular}
\end{table}

N-gram illustrates a device-dependent break-even point. With the same two-token lookup setting and 4-bit target, decode throughput changes by \(-8.9\%\) on M3 Ultra, whose paired baseline reaches \(162\) tokens/s, but by \(+13.0\%\) on M3 Max, whose baseline reaches \(92\) tokens/s. The \(-25.2\%\) result on the still slower M2 Pro confirms that baseline throughput alone is not predictive. Likewise, Classic on M4 Pro and Cascade-n-gram on M3 Max improve decode while remaining slower E2E, consistent with prefill offsetting decode-side savings.

DFlash and DDTree on M3 Pro produce the largest observed E2E gains, \(9.2\%\) and \(15.5\%\), respectively. Their paired BF16 baseline decodes at only \(15.89\) tokens/s, suggesting greater scope to amortize weight reads. A memory-bandwidth limitation is a plausible explanation: M2 Pro's higher nominal bandwidth (\(200\) versus \(150\) GB/s) coincides with a BF16 baseline near \(20\) tokens/s and a \(1.8\%\) DDTree decode loss. This comparison supports the hypothesis but does not isolate bandwidth from other processor and runtime differences. Gains also fail to persist across other tests: DFlash is \(15\%\) slower E2E on M4 Pro, and EAGLE-3 reduces M4 Pro decode throughput by \(16\%\).

\textbf{Conclusions.} Speculative decoding can accelerate particular Apple Silicon configurations, but the published server-GPU gains do not transfer robustly to the principal local chat benchmark. Its benefit depends jointly on accepted work, target-step cost, the complete draft--verify path, and prefill's share of latency. The positive cases fit this cost trade-off and reinforce the paper's broader conclusion that speedup belongs to a method--runtime--hardware--workload combination.

\subsection{Metal Kernels}
\label{sec:metal-kernels}


CUDA kernels exploit asynchronous copies, matrix units and cooperative grids
\cite{nvidia_cuda_2025}; MLX on Apple Silicon exposes a different contract
\cite{hannun_mlx_2023}. We test which mechanisms from Marlin
\cite{frantar_marlin_2024}, FlashInfer \cite{ye_flashinfer_2025} and grouped MoE
GEMM \cite{gale_megablocks_2023} survive in single-request decode. Each mechanism
is classified with the taxonomy of \S\ref{sec:taxonomy}, and matched RTX~4090
controls are used where possible (\S\ref{sec:nvidia}). A one-model megakernel
case study is reported separately in \S\ref{sec:mega}.

\subsubsection{The transfer taxonomy}
\label{sec:taxonomy}

A mechanism that fails to port does so for one of four reasons. Conflating them
is what makes porting work feel arbitrary; separating them makes the outcome
predictable. Every mechanism below carries exactly one label:

\begin{description}
  \item[\Transfers] The idea and its mechanism both exist on Metal. Port it.

  \item[\HwGap] The idea is sound but the hardware or language offers no
    equivalent instruction. \emph{Nothing you write closes this.} Example:
    \ccode{cp.async}.

  \item[\RuntimeGone] The overhead the technique removes has already been removed
    by the runtime, so the ceiling is near zero before you start. Example:
    kernel-launch amortisation under MLX's lazy graph.

  \item[\NotApplicable] The technique's precondition does not hold in our regime.
    Not a platform limitation at all --- the same code on the same hardware would
    need it if the regime changed. Example: sorting tokens by expert when the
    batch is one token.
\end{description}

The labels differ in who can act on them. \HwGap{} holds until the vendor ships
an instruction. \RuntimeGone{} is an instruction to measure the runtime before
writing a kernel. \NotApplicable{} marks a technique waiting on a workload ---
batching, speculative decoding, serving --- and so is deferred rather than
refuted. For each ported technique, the distribution of verdicts in its transfer
table --- not the speed-up the port produced --- is the result we argue from.

\subsubsection{Marlin: weight-only 4-bit GEMM}
\label{sec:marlin}

W4A16 stores weights as 4-bit codes and activations in \texttt{fp16}
\cite{frantar_gptq_2023}. Before multiplication, each code must be reconstructed
as a floating-point value. Materialising these values in memory would erase the
bandwidth saving (Listing~\ref{lst:naive}); Marlin instead reconstructs each
weight in registers, consumes it immediately, and never writes the expanded
representation to memory \cite{frantar_marlin_2024}.

\begin{cudacode}
// Strawman W4A16: correct, and it throws away the entire point.
dequantize<<<...>>>(w_q4, scales, biases, w_fp16);  // writes N*K fp16 to DRAM
gemm(x_fp16, w_fp16, out);                          // reads N*K fp16 from DRAM
\end{cudacode}
\captionof{listing}{The naive path. The GEMM reads \SI{16}{\bit} per weight, so
every byte saved by quantising is spent again --- the accuracy cost of 4-bit is
paid for no bandwidth gain.}
\label{lst:naive}

\paragraph{Mechanisms}
\label{sec:marlin:mechanisms}

\subparagraph{M1: cheap unpack.}
Marlin exploits the \texttt{fp16} bit layout: an \texttt{fp16} with the exponent
of $1024.0$ and a 4-bit integer $q$ in its low mantissa bits represents
$1024+q$. A mask and or construct this pattern, and NVIDIA's
\ccode{lop3.b32} performs the three-input Boolean operation in one instruction
\cite{nvidia_ptx_2025}. Eight nibbles fall from roughly 32 operations to about
eight (Listing~\ref{lst:lop3}). Shifted nibbles are not shifted down; their
power-of-two factor is folded into arithmetic that already occurs.

\begin{cudacode}
// Two nibbles at a time. 0x6400 is the fp16 bit pattern for 1024.0.
d = lop3(q & 0x000F000F, 0x64006400, LUT_AND_OR);  // -> {1024+q0, 1024+q1}
r = __hsub2(d, {1024.0h, 1024.0h});                // -> {q0, q1}, no cvt
\end{cudacode}
\captionof{listing}{Marlin's unpack. Eight nibbles go from roughly 32 operations
(shift, mask, convert, scale, per value) to about eight.}
\label{lst:lop3}

\subparagraph{M2: group factoring.}
For one scale and bias per group, the affine transform can be moved outside the
per-weight work:
\begin{equation}
  \sum_i (s\,q_i + b)\,x_i \;=\; s \sum_i q_i x_i \;+\; b \sum_i x_i .
  \label{eq:groupfactor}
\end{equation}
The scale is then applied once per group and no dequantised weight is formed.
Grouping itself is a checkpoint-format accuracy choice; this identity is what
makes that format cheap to consume.

\subparagraph{M3: asynchronous prefetch.}
\ccode{cp.async} and Hopper TMA copy the next tile from global to shared memory
while the current tile is multiplied \cite{nvidia_cuda_2025}, overlapping the
memory bus and matrix unit.

\subparagraph{M4: offline repacking.}
Weights and metadata are shuffled at load time into inner-loop order, producing
coalesced loads without runtime rearrangement.

\paragraph{What transfers}
\label{sec:marlin:transfer}

Register-only reconstruction, group factoring and repacking transfer; unpacking
needs different arithmetic. The matrix unit is \NotApplicable{} at $M=1$, and
public MSL makes asynchronous prefetch a \HwGap{}
(Table~\ref{tab:marlin-transfer}).

\begin{table}[htbp]
\centering
\small
\begin{tabular}{@{}llll@{}}
\toprule
\textbf{Mechanism} & \textbf{CUDA} & \textbf{Metal} & \textbf{Verdict} \\
\midrule
Never materialise \texttt{fp16} & registers  & registers                & \Transfers \\
M2 group factoring              & algebra    & algebra                  & \Transfers \\
M4 offline repacking            & load time  & \texttt{prepare\_model}  & \Transfers \\
M1 cheap unpack                 & \ccode{lop3} & prescale (Eq.~\ref{eq:prescale}) & \Transfers$^{\dagger}$ \\
Matrix unit                     & \ccode{mma}  & \mcode{simdgroup_matrix} & \NotApplicable$^{\ddagger}$ \\
M3 async prefetch               & \ccode{cp.async} & ---                & \HwGap \\
\bottomrule
\end{tabular}
\caption{Marlin transfer analysis. $^{\dagger}$~The idea transfers; the
instruction does not. Metal has no three-input logic op, so we reach the same end
by different arithmetic. $^{\ddagger}$~Metal \emph{has} a matrix unit, but at
$M=1$ there is no matrix to multiply --- this is a regime limitation, not a
platform one, and it reverses at $M \geq 2$ (\S\ref{sec:marlin:smallm}).}
\label{tab:marlin-transfer}
\end{table}

The asynchronous-prefetch verdict is narrower than it looks. The Apple GPU can
perform device-to-threadgroup asynchronous copies: private
\ccode{air.simdgroup_async_copy_1d/2d} intrinsics exist and Apple's MPS uses
them. They are undocumented and inaccessible from public MSL
\cite{apple_metal_2025}, making this an API gap rather than evidence of missing
silicon.

\paragraph{Metal implementation}
\label{sec:marlin:implementation}

Because MSL has no \ccode{lop3} equivalent \cite{apple_metal_2025}, the port
reaches M1 by exact power-of-two prescaling. For nibble $j$, Metal leaves the
value shifted in place and prescales its activation:
\begin{equation}
  \sum_j x_j\,q_j \;=\; \sum_j \bigl(x_j\,16^{-j}\bigr)\,\bigl(q_j \ll 4j\bigr).
  \label{eq:prescale}
\end{equation}
The factors are exact powers of two, preserving the result while removing
inner-loop shifts. The Metal path folds $16^{-j}$ into each activation rather
than folding the opposite factor into a scale; because activations are loaded
once per block of output rows, this amortises the adjustment. The final inner
loop combines group factoring, interleaved scale/bias metadata and activation
reuse, with $\Sigma x$ hoisted outside the row loop (Listing~\ref{lst:metal-qmv}).

\begin{metalcode}
// marlin_metal/csrc/marlin_qmv.metal
for (int i = 0; i < VPT; i += 4) {        // activations: loaded ONCE per block
    sum   += x0 + x1 + x2 + x3;           // @$\Sigma x$@ for the group-factored bias
    xt[i]     = x0;
    xt[i + 1] = x1 * (1.0f / 16.0f);      // fold @$16^{-j}$@ into x
    xt[i + 2] = x2 * (1.0f / 256.0f);
    xt[i + 3] = x3 * (1.0f / 4096.0f);
}
for (int r = 0; r < NR; r++) {            // NR output rows share that one load
    float accum = 0.0f;
    for (int i = 0; i < VPT / 4; i++) {
        accum += xt[4*i    ] * (wr[i] & 0x000f)   // masked in place: still
               + xt[4*i + 1] * (wr[i] & 0x00f0)   // shifted, and the prescale
               + xt[4*i + 2] * (wr[i] & 0x0f00)   // cancels the shift exactly
               + xt[4*i + 3] * (wr[i] & 0xf000);
    }
    const vec<T,2> s_b = sb_v[r * gpr + g];       // M4: one load, interleaved
    acc[r] = fma(s_b[0], accum, fma(s_b[1], sum, acc[r]));
}
\end{metalcode}
\captionof{listing}{The Metal inner loop. Three of Marlin's four mechanisms are
visible: group factoring in the final \mcode{fma}, interleaved metadata in the
single \mcode{vec<T,2>} load, and activation reuse across \mcode{NR} rows. Note
that $\Sigma x$ is hoisted out of the row loop --- the bias term of
Eq.~\eqref{eq:groupfactor} costs one add per element amortised over all
\mcode{NR} rows.}
\label{lst:metal-qmv}

Metal shaders are monomorphic, so the implementation instantiates variants for
three dtypes and six shape buckets:
\mcode{marlin_qmv_{f16,bf16,f32}_p{1,2,4}_r{4,8}}. Making \mcode{NR} a template
parameter rather than a runtime bound is worth \SI{10}{\percent}, because it
allows the compiler to unroll the row loop and allocate registers accordingly.

\paragraph{Results}
\label{sec:marlin:results}

At $M=1$ the kernel measures \x{0.95}--\x{1.00} against MLX \texttt{qmv} and
\x{0.993} end to end. MLX already reaches \gbs{440}, about
\SI{80}{\percent} of DRAM peak, leaving little room for cheaper unpacking; a
6-bit \texttt{lm\_head} matvec likewise measured parity, because MLX runs that
shape at the same rate at 4, 6 and 8 bits (App.~\ref{app:marlin:lmhead}).
\label{sec:marlin:smallm}
The picture changes at $M=2$--$8$. Stock MLX's tiled GEMM reaches roughly \x{5}
the $M=1$ time at $M=8$ despite identical weight traffic. Dequantising each block
once into registers and reusing it across activation rows
(Listing~\ref{lst:metal-qmm}) changes $t(8)/t(1)$ to $4.1$--$4.5$, versus MLX's
$4.9$--$5.1$, and yields \x{1.09}--\x{1.20} on all tested Qwen3-4B and Qwen3-8B
transformer-layer shapes \cite{yang_qwen3_2025}.

\begin{metalcode}
// Fully dequantise this block into registers ONCE, folding s and b in, so the
// per-m loop below is pure load+FMA and the dequant amortises over all M rows.
float wf[NR][VPT];
for (int r = 0; r < NR; r++) {
    const vec<T,2> s_b = sb_v[r * gpr + g];
    const float s = float(s_b[0]), b = float(s_b[1]);
    for (int i = 0; i < 4; i++) {
        const uint16_t wv = w_row[i];
        wf[r][4*i    ] = fma(s,                 float(wv & 0x000f), b);
        wf[r][4*i + 1] = fma(s * (1.0f/16.0f),  float(wv & 0x00f0), b);
        wf[r][4*i + 2] = fma(s * (1.0f/256.0f), float(wv & 0x0f00), b);
        wf[r][4*i + 3] = fma(s * (1.0f/4096.0f),float(wv & 0xf000), b);
    }
}
for (int m = 0; m < NM; m++) { /* pure load + FMA against wf[][] */ }
\end{metalcode}
\captionof{listing}{Small-$M$ path. Here the weights \emph{are} widened into
registers, because they will be reused $M$ times --- the opposite trade from
Listing~\ref{lst:metal-qmv}, and correct for the same reason.}
\label{lst:metal-qmm}

CUDA's stock 4-bit kernel already serves $M=8$ for only
\SIrange{5}{12}{\percent} more time than $M=1$ (\S\ref{sec:nvidia:w4a16}), so the
Apple gain closes an MLX runtime gap rather than recovering an instruction-level
mechanism.

\paragraph{Speculative decoding}
\label{sec:marlin:spec}

The verify pass of speculative decoding naturally produces $M\in[2,8]$
\cite{leviathan2023fast}, making the small-$M$ path directly useful.
Table~\ref{tab:marlin-spec} reports Qwen3-8B-4bit with a Qwen3-0.6B-4bit draft.

\begin{table}[htbp]
\centering
\small
\begin{tabular}{@{}rrrrr@{}}
\toprule
$k$ & verify $M$ & stock (tok/s) & marlin (tok/s) & ratio \\
\midrule
2 & 3 & 78.5 & 79.9 & \x{1.02} \\
4 & 5 & 68.6 & 72.9 & \x{1.06} \\
7 & 8 & 51.9 & \textbf{59.0} & \x{\mathbf{1.14}} \\
8 & 9 & 45.1 & 45.1 & \x{1.00} \\
\bottomrule
\end{tabular}
\caption{Speculative decoding, 256 tokens, median of 2. Plain decode is
52.4~tok/s. At $k=7$ speculation is break-even with stock kernels
(\x{0.99} against plain decode) and becomes \x{1.13} with ours. At $k=8$ the
verify is $M=9$; our kernel disengages and both paths measure \x{1.00}.}
\label{tab:marlin-spec}
\end{table}

\subsubsection{FlashInfer: decode attention}
\label{sec:fi}

FlashInfer combines online softmax \cite{milakov_online_2018,dao_flashattention_2022}
with split-KV, which partitions the cache and merges partials to restore
batch-1 parallelism \cite{dao_flashdecoding_2023}.

\paragraph{Mechanisms and constraints}
\label{sec:fi:mechanisms}

Decode attention has two awkward properties. First, its softmax cannot be
materialised cheaply: stabilising it requires the maximum score, but storing all
$N$ scores would write and reread a vector per head. Online softmax (M1) instead
maintains $(m,\ell,o)$ and rebases the accumulated history when $m$ changes:
\begin{align}
  m' &= \max(m, s), \qquad \alpha = e^{\,m - m'} \label{eq:online-alpha}\\
  \ell &\leftarrow \alpha\,\ell + e^{\,s - m'}, \qquad
  o \leftarrow \alpha\,o + e^{\,s - m'}\,v .
  \label{eq:online-update}
\end{align}
This gives one pass and memory constant in $N$.

Second, decode removes the positions dimension from available parallelism. A
single request with 32 heads exposes only 32 large serial units. Split-KV (M2)
partitions the cache into $C$ chunks, turning these into $32C$ independent units;
each returns a local $(o_c,\ell_c,m_c)$ merged by
\begin{equation}
  m = \max_c m_c, \qquad
  o = \sum_c o_c\,e^{\,m_c-m}, \qquad
  \ell = \sum_c \ell_c\,e^{\,m_c-m}, \qquad
  \mathrm{out}=o/\ell .
  \label{eq:merge}
\end{equation}
The same rescaling identity makes the loop both streamable and composable.

FlashInfer adds two serving-oriented mechanisms. A paged, block-sparse KV cache
(M3) lets one template address paged caches \cite{kwon_efficient_2023}, ragged
batches and shared prefixes. A host-side plan (M4) distributes chunks from
unequal sequence lengths so no CTA idles. Listing~\ref{lst:fi-cuda} shows the
resulting decode loop.

\begin{cudacode}
plan = scheduler(seq_lens, num_heads);      // host: balance ragged batches
for (tile in my_chunk) {                    // each CTA owns one (head, chunk)
    cp.async(smem[nxt], paged_kv[tile + 1]);
    s  = qk_mma(q, smem[cur]);              // matrix unit
    m2 = max(m, rowmax(s));                 // online max update
    o  = o * exp(m - m2) + pv_mma(exp(s - m2), v);
    l  = l * exp(m - m2) + rowsum(exp(s - m2));
    m  = m2;
}
write_partial(o, m, l);                     // consumed by the merge kernel
\end{cudacode}
\captionof{listing}{FlashInfer's decode loop, abridged. Three of its four
mechanisms are visible in the first two lines and the last.}
\label{lst:fi-cuda}

\paragraph{What transfers}
\label{sec:fi:transfer}

Online softmax, split-KV and zero-copy cache views transfer. Paging and
load-balanced planning require concurrent serving and are \NotApplicable{};
asynchronous prefetch is a \HwGap{} (Table~\ref{tab:fi-transfer}).

\begin{table}[htbp]
\centering
\small
\begin{tabular}{@{}llll@{}}
\toprule
\textbf{Mechanism} & \textbf{CUDA} & \textbf{Metal} & \textbf{Verdict} \\
\midrule
M1 online softmax     & running triple            & identical            & \Transfers \\
M2 split-KV           & 2 kernels + workspace     & 2 dispatches         & \Transfers \\
Zero-copy cache view  & pointer arithmetic        & strides at eval time & \Transfers \\
M3 paged / block-sparse & page table              & ---                  & \NotApplicable \\
M4 load-balanced plan & host scheduler            & lookup table         & \NotApplicable \\
$QK^{\!\top}$ on matrix unit & \ccode{mma}        & \mcode{simd_sum}     & \NotApplicable \\
Async prefetch        & \ccode{cp.async}          & ---                  & \HwGap \\
\bottomrule
\end{tabular}
\caption{FlashInfer transfer analysis. Unlike Marlin
(Table~\ref{tab:marlin-transfer}), most of what does not transfer is
\NotApplicable{} rather than \HwGap: MLX's KV cache is a single growing
contiguous slice, so there are no pages to index, and at batch~1 there is nothing
to load-balance.}
\label{tab:fi-transfer}
\end{table}

\paragraph{Metal and runtime implementation}
\label{sec:fi:implementation}

The online-softmax equations transcribe unchanged (Listing~\ref{lst:fi-metal}).
Metal uses \mcode{simd_sum}, a cross-lane shuffle touching no memory, instead of
a matrix unit for $QK^{\!\top}$. MLX presents its growing cache as a strided
view; materialising that view per call would dominate attention, so the
primitive reads strides at evaluation time (\mcode{ensure_row_contiguous=false}).
This is not a page table, but supplies the required property here: addressing
the cache without copying it.

\begin{metalcode}
// cider_mlx/csrc/cider_sdpa_vector.metal
for (int i = simd_gid; i < N; i += BN) {          // simdgroups stride the cache
    U score = 0;
    for (int j = 0; j < qk_per_thread; j++) score += q_reg[j] * k_ptr[j];
    score = simd_sum(score);                      // no matrix unit: 32-lane reduce

    U new_max   = max(max_score, score);          // online max update
    U factor    = fast::exp(max_score - new_max);
    U exp_score = fast::exp(score - new_max);
    max_score     = new_max;                      // online accumulator update
    sum_exp_score = sum_exp_score * factor + exp_score;
    for (int j = 0; j < v_per_thread; j++)
        o_reg[j] = o_reg[j] * factor + exp_score * U(v_ptr[j]);

    k_ptr += inner_k_stride;                      // strides read at eval time:
    v_ptr += inner_v_stride;                      // the cache slice stays zero-copy
}
\end{metalcode}
\captionof{listing}{The Metal kernel. The online-softmax update transcribes
unchanged. The dot product is reduced with \mcode{simd_sum}, a cross-lane shuffle
that touches no memory, in place of the matrix unit.}
\label{lst:fi-metal}

Runtime integration matters independently of shader quality. A Python wrapper
around the identical shader measured \x{0.63} against stock at $N=256$,
\x{0.73} at $N=1024$, and became favourable only at $N=16384$ (\x{1.34}). As a
native MLX primitive the same source measures \x{1.06}, \x{1.06} and \x{1.38}.
The kernel was not responsible for the short-context loss; per-call wrapper
overhead was.

\paragraph{Results}
\label{sec:fi:results}

\begin{figure}[t]
\centering
\includegraphics[width=\linewidth]{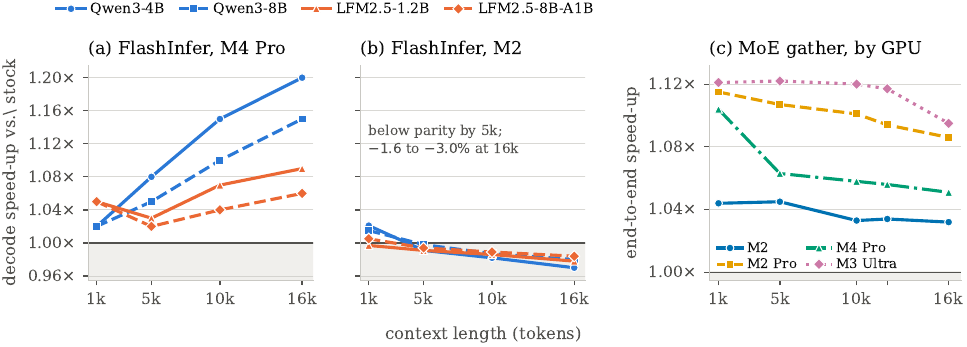}
\caption{Decode-throughput speed-up across context lengths. FlashInfer reaches
\x{1.20} on M4~Pro but \x{0.97} on M2 (a,b), whereas MoE gather provides
consistent gains across all four Apple GPUs (c). Exact values are reported in
Tables~\ref{tab:fi-context} and~\ref{tab:moe-compose}.}
\label{fig:kernel-results}
\end{figure}

End to end, the gain reaches \x{1.20} on M4~Pro but falls to \x{0.97} on M2
(Fig.~\ref{fig:kernel-results}a,b; Table~\ref{tab:fi-context}).

\begin{table}[htbp]
\centering
\small
\begin{tabular}{@{}lrrrr@{}}
\toprule
& \multicolumn{4}{c}{speed-up vs.\ stock, by context length} \\
\cmidrule(l){2-5}
\textbf{Model} & 1k & 5k & 10k & 16k \\
\midrule
\multicolumn{5}{@{}l}{\emph{M4~Pro}} \\
Qwen3-4B (dense)        & \x{1.02} & \x{1.08} & \x{1.15} & \x{1.20} \\
Qwen3-8B (dense)        & \x{1.02} & \x{1.05} & \x{1.10} & \x{1.15} \\
LFM2.5-1.2B (hybrid)    & \x{1.05} & \x{1.03} & \x{1.07} & \x{1.09} \\
LFM2.5-8B-A1B (hybrid)  & \x{1.05} & \x{1.02} & \x{1.04} & \x{1.06} \\
\midrule
\multicolumn{5}{@{}l}{\emph{M2}} \\
Qwen3-4B (dense)        & \x{1.021} & \x{0.991} & \x{0.982} & \x{0.970} \\
Qwen3-8B (dense)        & \x{1.015} & \x{0.998} & \x{0.987} & \x{0.980} \\
LFM2.5-1.2B (hybrid)    & \x{0.997} & \x{0.991} & \x{0.986} & \x{0.978} \\
LFM2.5-8B-A1B (hybrid)  & \x{1.005} & \x{0.994} & \x{0.989} & \x{0.984} \\
\bottomrule
\end{tabular}
\caption{End-to-end decode, generation length 256 --- the per-cell figures behind
Fig.~\ref{fig:kernel-results}. M2 is reported to three decimals because its whole
range is within \SI{3}{\percent} of parity; at two decimals LFM2.5-1.2B at 1k
rounds to \x{1.00} and its \SI{0.3}{\percent} loss disappears.}
\label{tab:fi-context}
\end{table}

The end-to-end ordering follows model structure. Dense Qwen3-4B moves most;
Qwen3-8B moves less because larger MLP and projection matmuls dilute attention;
hybrid LFM2.5-8B-A1B caps at \x{1.06} because only 6 of 24 layers use attention.
The fraction of token time touched by a kernel bounds its effect in either
direction.

\label{sec:fi:machines}
In isolation the attention kernel changes sign across Apple GPU families
(Table~\ref{tab:fi-machines}).

\begin{table}[htbp]
\centering
\small
\begin{tabular}{@{}lrrr@{}}
\toprule
\textbf{Machine} & ctx 1k & ctx 16k & dependent-chain bubble \\
\midrule
M2        & $+0.4\%$ & $-1.7\%$ & \SI{7.0}{\percent} \\
M2~Pro    & $-0.5\%$ & $-5.7\%$ & \SI{24.0}{\percent} \\
M4~Pro    & $+2.3\%$ & $+6.4\%$ & \SI{32.7}{\percent} \\
M3~Ultra  & $+1.4\%$ & $+2.2\%$ & \SI{39.9}{\percent} \\
\bottomrule
\end{tabular}
\caption{Attention kernel alone on LFM2.5-8B-A1B, generation length 512. The
serialisation bubble is the shortfall between dependent and independent chains
of 24 matvecs: bandwidth lost while one dispatch drains and the next ramps.}
\label{tab:fi-machines}
\end{table}

On M2 and M2~Pro the split-KV loss grows with context, the mirror image of the
large-GPU curve. Prefill is unchanged on M2 (\x{1.00} across models and
contexts, TTFT within \SI{0.5}{\percent}), locating the effect in decode. The
cause is split-KV itself: a second dispatch and a partials round trip have fixed
cost, while extra parallelism has value only when one query does not already
occupy the GPU. RTX~4090/FlashInfer and Metal/MLX also trend oppositely with
context, but different baselines prevent hardware-only attribution
(\S\ref{sec:nvidia:attn}).

Split-KV is therefore enabled per architecture from a measured table, and an
unmeasured GPU keeps stock MLX routing; the flat $N\geq1024$ threshold of the
reference port cost $1$--$3\%$ on M2 (App.~\ref{app:fi:routing}).

\subsubsection{MoE gather kernel}
\label{sec:moe}

An MoE router selects the top-$k$ experts for each token, after which the gather
applies a different weight matrix to each selected slot. NVIDIA grouped-GEMM
and block-sparse kernels organize many tokens per expert
\cite{nvidia_cutlass_2024,gale_megablocks_2023,kwon_efficient_2023}, unlike
batch-1 decode.

\paragraph{Mechanisms and their preconditions}
\label{sec:moe:mechanisms}

\subparagraph{M1: sort by expert.}
Permute token rows so that all rows routed to one expert are contiguous; each
expert's slice then becomes an ordinary dense GEMM. The irregularity is paid
once instead of inside the inner loop.

\subparagraph{M2: grouped GEMM.}
One launch performs $E$ matmuls with distinct shapes and base pointers. A
problem visitor walks the $(M_e,N,K,A_e,B_e)$ tuples and distributes tiles
across all groups, preventing an expert with few tokens from leaving a
multiprocessor idle.

\subparagraph{M3: block-sparse rather than padded.}
MegaBlocks avoids rounding every expert's token count to a tile multiple by
expressing the layer as one block-sparse matmul with variable block sizes. This
avoids both wasted padded compute and quality loss from dropping overflow.

\subparagraph{M4: fuse the block.}
Gate, up, the SwiGLU nonlinearity \cite{shazeer_glu_2020}, down and weighted
reduction are combined in as few launches as possible, keeping large
intermediates on chip.

\paragraph{What transfers}
\label{sec:moe:transfer}

All four mechanisms assume many rows per expert. At batch~1 there is nothing to
sort, no group larger than one row, no tile quantisation to avoid and no
load-balancing problem. Sorting, grouped scheduling, block sparsity and block
fusion are therefore \NotApplicable{}; asynchronous prefetch is a \HwGap{}
(Table~\ref{tab:moe-transfer}).

\begin{table}[htbp]
\centering
\small
\begin{tabular}{@{}llll@{}}
\toprule
\textbf{Mechanism} & \textbf{Precondition} & \textbf{At batch 1} & \textbf{Verdict} \\
\midrule
M1 sort by expert    & many rows per expert    & one row  & \NotApplicable \\
M2 grouped GEMM      & groups worth scheduling & $M_e = 1$ & \NotApplicable \\
M3 block-sparse      & tile-size quantisation  & no tiles & \NotApplicable \\
M4 fuse the block    & large intermediates     & one vector & \NotApplicable$^{\dagger}$ \\
Matrix unit          & $M \geq$ tile height    & $M = 1$  & \NotApplicable \\
Async prefetch       & \ccode{cp.async}        & ---      & \HwGap \\
\bottomrule
\end{tabular}
\caption{MoE grouped GEMM transfer analysis. Every mechanism is
\NotApplicable{} at batch~1 --- an unusually clean case.
$^{\dagger}$~We tested M4 anyway; see \S\ref{sec:moe:followups}.}
\label{tab:moe-transfer}
\end{table}

\paragraph{Metal implementation}
\label{sec:moe:metal}

The implementation therefore reduces to the same quantized inner loop with one
expert-weight indirection. For router indices $\iota_j$, the batch-1 operation is
\begin{equation}
  \mathrm{out}[j] \;=\; x \cdot
  \mathrm{dequant}\bigl(W[\,\iota_j\,]\bigr)^{\!\top},
  \qquad j=1\dots k .
  \label{eq:gather-qmv}
\end{equation}
It differs from the dense matvec only in the weight base address.

Three details distinguish the MoE gather from the dense matvec.
\texttt{mlx-lm}'s \texttt{SwitchGLU} uses \emph{broadcast} inputs for gate and up
(one activation shared across the token's $k$ slots) and \emph{per-slot} inputs
for down (one vector per expert); one flag selects the address calculation.
LFM2.5's expert width $K=1792$ divides no convenient tiling, so a wide main pass
covers the largest tiled prefix and a guarded single-word loop handles the tail.
Finally, the offline scale/bias interleaving of \S\ref{sec:marlin} transfers
unchanged. The resulting two pointer offsets appear in
Listing~\ref{lst:moe-metal}.

\begin{metalcode}
// moe_metal/csrc/gather_qmv.metal -- the entire MoE-specific part
const int slot   = int(tid.y);              // which of the token's top-k slots
const int expert = int(indices[slot]);      // expert gather, one load
const int m      = per_slot ? slot : slot / topk;

const device T* x_ptr = x + size_t(m) * K;
const device uint16_t* ws = (const device uint16_t*)
    (w + (size_t(expert) * N + row0) * K_w32);        // offset the weights
const device vec<T,2>* sb_v = (const device vec<T,2>*)sb
    + (size_t(expert) * N + row0) * gpr;              // and the metadata

// from here it is byte-for-byte the dense matvec chassis
gather_qmv_body<T, BITS, ROWS>(x_ptr, ws, sb_v, K, gpr, gs_shift, ...);
\end{metalcode}
\captionof{listing}{Two pointer offsets. The expert identity enters the kernel as
arithmetic on a base address and nothing else --- no sort, no permutation, no
scatter, and no divergent branch, since \mcode{slot} is uniform across the
threadgroup.}
\label{lst:moe-metal}

\paragraph{Results and composition with attention}
\label{sec:moe:composition}

Against \texttt{mx.gather\_qmm}, the kernel reaches \x{1.05} on gate/up and
\x{1.23} on down; patching 66 \texttt{SwitchLinear} layers yields
\x{1.03}--\x{1.12} end to end across machines (Fig.~\ref{fig:kernel-results}c):

\begin{center}\small
\begin{tabular}{@{}lr@{}}
\toprule
\textbf{Machine} & \textbf{Speed-up range} \\
\midrule
M2       & \x{1.032}--\x{1.045} \\
M2~Pro   & \x{1.086}--\x{1.115} \\
M4~Pro   & \x{1.051}--\x{1.104} \\
M3~Ultra & \x{1.095}--\x{1.122} \\
\bottomrule
\end{tabular}
\end{center}

\noindent
The gain comes from removing machinery batch-1 decode does not need, not from
recreating grouped GEMM. Expert-weight traffic is independent of KV-cache length,
whereas attention traffic grows with it. The two kernels therefore affect
disjoint operations, the MoE contribution stays roughly flat with context, and
the gains compose almost multiplicatively (Table~\ref{tab:moe-compose}).

\begin{table}[htbp]
\centering
\small
\begin{tabular}{@{}rrrrrr@{}}
\toprule
ctx & stock (\si{\tok\per\second}) & attention & MoE & both & product \\
\midrule
1000  & 167.3 & \x{1.012} & \x{1.068} & \x{1.075} & \x{1.081} \\
5000  & 158.5 & \x{1.024} & \x{1.060} & \x{1.090} & \x{1.085} \\
10000 & 149.4 & \x{1.042} & \x{1.058} & \x{1.106} & \x{1.102} \\
12000 & 145.6 & \x{1.051} & \x{1.058} & \x{1.113} & \x{1.112} \\
16000 & 140.0 & \x{1.064} & \x{1.052} & \x{1.114} & \x{1.119} \\
\bottomrule
\end{tabular}
\caption{Composition on M4~Pro. The final column is the product of the two
single-kernel columns. Disjoint kernels multiply to within measurement noise.}
\label{tab:moe-compose}
\end{table}

\paragraph{Follow-up experiments}
\label{sec:moe:followups}

\subparagraph{Down projection plus routing reduction (kept).}
Combining the down matvec, router-score multiply and top-$k$ sum removes a
genuine intermediate tensor and adds $1.2$--$3.0\%$ on top of the gather.

\subparagraph{Gate, up and SwiGLU fusion (not enabled).}
This measured \x{0.986}--\x{0.999} on LFM2.5's coarse experts. It reads the same
bytes as the separate kernels while adding register pressure.

\subparagraph{Explicit repeated-expert deduplication (negative).}
Although two slots selecting the same expert appear to read its rows twice,
explicit staging measured only \x{0.76}--\x{0.93}. Apple's system-level cache
already serviced the repeat, leaving the staged version to pay ownership
imbalance and register pressure for a saving already made.

\subparagraph{Grouped GEMM in prefill (open negative).}
The grouped-GEMM precondition does hold during prefill. Our bit-exact
implementation nevertheless measures \x{0.47} against stock on M4~Pro. This is
an open implementation result, distinct from the \NotApplicable{} verdict for
batch-1 decode.

\subsubsection{Exploratory megakernel case study}
\label{sec:mega}

A megakernel fuses an entire block or forward pass into one persistent launch:
the grid remains resident, phases exchange data through memory, and later loads
can be pipelined behind earlier computation. This can remove launch boundaries
and inter-kernel bubbles, and has produced strong batch-1 results on NVIDIA
systems \cite{noauthor_look_nodate,wu_mirage_2024}.

We treat the following experiment as an exploratory case study, not as one of
the primary porting results. It covers one LFM2.5-8B-A1B MoE block on one
M4~Pro under MLX, and the implementation was not evaluated broadly enough to
support a claim about every model, Apple GPU or possible megakernel design. Its
value is diagnostic: it exposes several conditions that a persistent,
grid-synchronised design would have to overcome on public Metal.

Our hypothesis is that this class of megakernel is poorly matched to the current
Metal/MLX contract for three reasons. MLX already amortises most dispatch and CPU
encoding overhead; public Metal provides no cooperative-launch, grid-barrier or
occupancy-query contract; and one persistent grid forces heterogeneous phases to
share geometry and carry their combined resource footprint. The first two are
properties of the exposed runtime/API contract. The size of the geometry penalty
is specific to the model and kernel tested here.

The prototype fused eleven operations --- router, fp32 softmax, top-$k$, gated
experts, down projection and weighted reduction --- into one bit-faithful
dispatch. With the barrier disabled to isolate compute, it measured \us{174}
against \us{96} for the stock chain, a \x{1.8} slowdown. This observation does
not establish that megakernels cannot work on Metal; it motivates the
precondition tests and interpretation below.

\paragraph{P1: what a dispatch boundary actually costs}
\label{sec:mega:p1}

\begin{center}\small
\begin{tabular}{@{}lr@{}}
\toprule
GPU-side dispatch boundary            & \us{3.04} \\
CPU graph build and encode per token  & \SI{0.54}{\milli\second} \\
\dots of which visible in wall-clock  & $\approx 0$ \\
\bottomrule
\end{tabular}
\end{center}

\noindent
The last row is the one that matters: a pipelined token takes
\SI{6.04}{\milli\second} against \SI{6.53}{\milli\second} for a synchronous
\texttt{eval}, so the \SI{0.54}{\milli\second} of graph building is already
hidden. MLX encodes a token's whole graph into a few command buffers with
consecutive dispatches sharing an encoder, and \texttt{mlx-lm} overlaps token
$t{+}1$'s CPU work with token $t$'s GPU work; on NVIDIA, by contrast, a
\SIrange{3}{10}{\micro\second} launch against a kernel of a few microseconds puts
boundaries at up to half of wall-clock. Summed over the MoE block the boundaries
total \us{15} --- the entire upside of fusing it.

\paragraph{Cooperation, occupancy and geometry}
\label{sec:mega:probes}

Three further probes, detailed in App.~\ref{app:mega}, show why the fused grid
cannot be trusted or made fast. Public Metal offers no cooperative launch: a
hand-rolled grid barrier over relaxed atomics works, but plain device stores
were \SI{96}{\percent} stale across threadgroups, and a register-heavier variant
of the same kernel stopped converging with no occupancy query to predict it
(App.~\ref{app:mega:p2}--\ref{app:mega:p3}). Forcing all phases onto one grid
also cost \us{78} on this block, about \x{5} the \us{15} that removing every
boundary can recover (App.~\ref{app:mega:geometry}).

\paragraph{Where the dependency bubble actually goes}
\label{sec:mega:bubble}

The dependency bubble a megakernel targets is real: 24 dependent matvecs run at
\gbs{77} where the same 24 run independently at \gbs{114}, a
\SI{32.7}{\percent} serialisation loss on M4~Pro, and token accounting leaves
\SI{2}{\milli\second} of a \SI{6.04}{\milli\second} token unexplained by summed
operation times. The pool exists; the question is what reaches it
(Table~\ref{tab:mega-levers}).

\begin{table}[htbp]
\centering
\small
\begin{tabular}{@{}lll@{}}
\toprule
\textbf{Experiment} & \textbf{Mechanism} & \textbf{End-to-end} \\
\midrule
Fused add + RMSNorm      & fewer bytes (one pass)         & $+1$ to $+1.8\%$ \\
MoE gather, $M \geq 2$   & better rate                    & \x{1.25} at $M{=}8$ \\
Conv mixer, 6 ops to 1   & \x{2.2} rate, 1 dispatch       & \x{1.01}--\x{1.07}, short ctx only \\
4-bit KV cache           & fewer KV bytes                 & \x{1.17}--\x{1.37} below 2k \\
Explicit expert dedup    & remove repeated reads          & \textbf{negative} \\
6-bit \texttt{lm\_head}  & cheaper unpack                 & parity (rate-capped shape) \\
Whole MoE block fused    & remove all dispatches          & \textbf{negative} \\
\bottomrule
\end{tabular}
\caption{Seven attempts on the same bubble. Every positive result moved fewer
bytes or moved them faster; attempts aimed only at removing dispatches were
neutral or negative.}
\label{tab:mega-levers}
\end{table}

In this Apple/MLX configuration, dispatch boundaries are nearly free and byte
traffic dominates. Decode runs at roughly \gbs{170} effective against
per-operation rates of \gbs{250}--\gbs{440}; these experiments reached the gap
by shortening dependent chains --- speculative decoding or batching --- or by
shrinking the byte budget, not by replacing dispatches with a persistent grid.

\paragraph{Interpretation and limitations}
\label{sec:mega:interpretation}

In our view, persistent multi-phase megakernels are unlikely to be a robust
optimisation under the current public Metal/MLX contract when they require
grid-wide barriers. The removable boundary cost was only \us{15} in this case,
whereas the fixed-grid penalty was \us{78}; correctness also depended on
co-residency that Metal does not guarantee or expose through an occupancy query.
These conditions make performance and forward progress sensitive to compiler
resource allocation. An interpreter-style design \cite{wu_mirage_2024} would
retain both pressures (App.~\ref{app:mega:interpreter}).

This is a scoped interpretation, not an impossibility result. A narrower fused
kernel without grid-wide cooperation, a different model geometry, a future
Metal API, or a runtime with larger visible dispatch costs could change the
trade. The transferable recommendation is to measure boundary cost, cooperation
guarantees, occupancy sensitivity and per-phase geometry before committing to a
persistent implementation.

\subsubsection{RTX 4090 comparison}
\label{sec:nvidia}

The RTX~4090 controls below measure two platform-level quantities used by the
kernel analysis: the gap between dependent and independent matvec chains, and
the cost of a dispatch boundary. The matched runs use PyTorch~2.13, CUDA~13.2,
the same chain length, shape, quantisation group and interstitial operation as
the Apple experiment. They provide comparative evidence for these two
mechanisms, not a direct counterpart for every kernel verdict.

\paragraph{Method and pitfalls}
\label{sec:nvidia:method}

Kernel timings are medians of 50 CUDA-event-timed repetitions after warm-up;
the dispatch experiment uses 20 repetitions. Three details changed the numbers
by more than the effects being measured, and are recorded because a reader
reproducing this will hit them:

\begin{itemize}
  \item \textbf{L2 residency.} The 4090 holds \SI{75}{\mega\byte} of L2, which
    swallows a whole layer's weights (\SI{49.8}{\mega\byte} at
    $K{=}2560,N{=}9728$ in \texttt{fp16}) and a 16k KV cache
    (\SI{67}{\mega\byte}) alike. Timing one buffer repeatedly therefore reports
    cache bandwidth: our first attempt produced \gbs{1277} and \gbs{1337},
    both above the card's \gbs{1008} DRAM peak. Every bandwidth measurement
    below rotates over enough distinct buffers to exceed L2 by $3\times$.
  \item \textbf{GQA expansion.} Materialising the grouped-query KV cache with
    \texttt{repeat\_interleave} makes the baseline read $h_q/h_{kv}=4$ times more
    bytes than the kernel under test, inflating the ratio fourfold. SDPA is
    therefore called with \texttt{enable\_gqa=True}.
  \item \textbf{Implicit batching.} \texttt{llm.generate(prompts)} runs the whole
    prompt set concurrently; the speculative sweep uses
    \texttt{max\_num\_seqs=1} so that it measures single-request decode rather
    than batch-32 throughput, where speculation loses for unrelated reasons.
\end{itemize}

The matched contract experiment and the reported PyTorch int4 measurements use
group size 64, as on Apple. PyTorch's int4 primitive requires BF16 activations,
where the Apple path uses FP16. The withheld vLLM Marlin attempt used group
size 128.

\paragraph{The platform contract}
\label{sec:nvidia:contract}

\begin{figure}[htbp]
\centering
\includegraphics[width=0.80\linewidth]{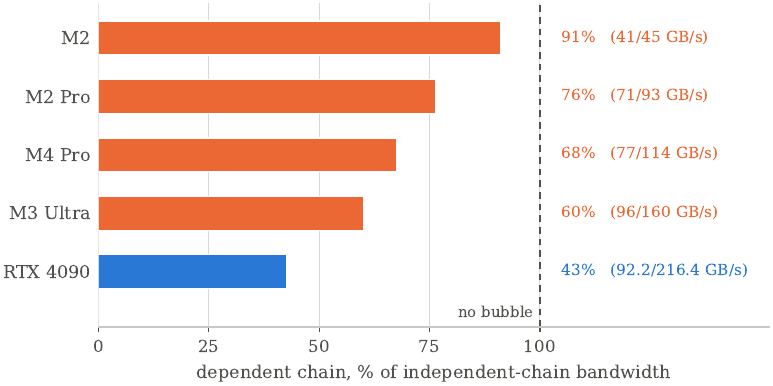}
\caption{Dependent-chain bandwidth as a share of the independent chain. All
runs use 24 square $2048\times2048$ 4-bit matvecs with group size 64; the
dependent path applies the same interstitial multiply on both platforms.}
\label{fig:platform-contract}
\end{figure}

\begin{table}[htbp]
\centering
\small
\begin{tabular}{@{}llrrrr@{}}
\toprule
\textbf{Shape} & \textbf{Weights} & \textbf{dependent} & \textbf{independent}
  & \textbf{penalty} & \textbf{of peak} \\
\midrule
$K{=}N{=}2048$ & 4-bit, group 64 & \gbs{92.2} & \gbs{216.4} & \SI{57.4}{\percent} & \SI{21.5}{\percent} \\
\bottomrule
\end{tabular}
\caption{Matched RTX~4090 chain with BF16 activations. The penalty exceeds the
Apple range of \SI{7.0}{\percent}--\SI{39.9}{\percent}, while the independent
chain reaches \SI{21.5}{\percent} of advertised DRAM bandwidth.}
\label{tab:nvidia-contract}
\end{table}

The matched result reverses the earlier layer-shape measurement: the RTX~4090
loses \SI{57.4}{\percent}, more than any tested Apple GPU. This quantity includes
the interstitial elementwise dispatches in the dependent path, exactly as in the
Apple experiment, so we interpret it as a dependency-chain penalty rather than
a pure matvec roofline effect. Issuing the independent matvecs on 24 CUDA streams
falls further to \gbs{65.3} (\x{0.302} against one stream), so naive concurrency
does not recover the gap.

\paragraph{What a dispatch boundary costs}
\label{sec:nvidia:dispatch}

\begin{figure}[htbp]
\centering
\includegraphics[width=0.72\linewidth]{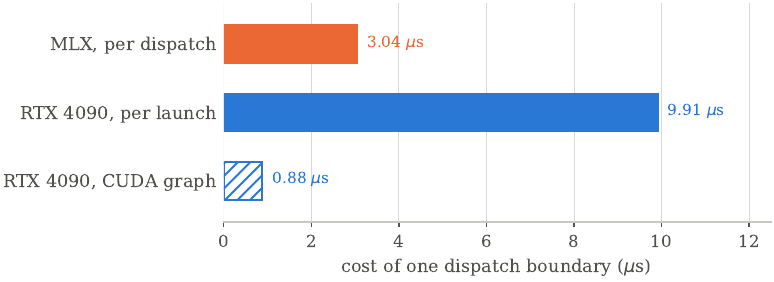}
\caption{Per-boundary cost. CUDA graph capture is hatched because it is a
different mechanism, not a peer measurement: it removes boundaries rather than
paying for them.}
\label{fig:dispatch-boundary}
\end{figure}

The matched chain of 512 dependent FP16 multiplies over 2048 elements takes
\SI{5.07}{\milli\second}, or \us{9.91} per launch --- \x{3.26} MLX's \us{3.04}
(\S\ref{sec:mega:p1}). Captured as a CUDA graph the same chain takes
\SI{0.45}{\milli\second} (\us{0.88} per op), recovering
\SI{91.2}{\percent} of the boundary cost.

This reverses the usual framing. NVIDIA has more boundary cost to reclaim than
Apple does \emph{and} a supported mechanism for reclaiming most of it without a
persistent kernel; MLX has already amortised its own boundaries through command
encoding and offers no equivalent capture. A megakernel on NVIDIA competes with
graph capture for the same \SI{91}{\percent}; on Apple it competes for \us{15}
per block that MLX has largely taken already.

\paragraph{W4A16 by batch size}
\label{sec:nvidia:w4a16}

\begin{figure}[htbp]
\centering
\includegraphics[width=0.80\linewidth]{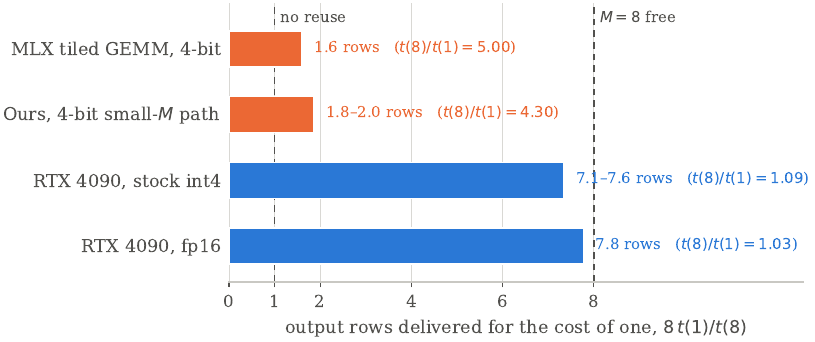}
\caption{Output rows delivered for the time one row costs, $8\,t(1)/t(8)$, for a
4-bit linear at MLP width: eight means a batch of eight is free, one means no
weight reuse at all. Apple ranges span the transformer layer shapes of
\S\ref{sec:marlin:results}; the 4090 range spans the two reported Qwen3-4B and
Qwen3-8B widths. The 4090's figure is if anything flattered by its own overhead:
its int4 kernel sustains only \gbs{507} at $M{=}1$ where \texttt{fp16} on the same
shape reaches \gbs{800}, so roughly \us{10} of the \us{24.6} baseline is fixed
cost, which inflates $t(1)$ and shrinks the ratio.}
\label{fig:m-scaling}
\end{figure}

\begin{table}[htbp]
\centering
\small
\begin{tabular}{@{}lrrrrr@{}}
\toprule
\textbf{Shape / kernel} & $M{=}1$ & $M{=}2$ & $M{=}4$ & $M{=}8$ & $t(8)/t(1)$ \\
\midrule
$K{=}2560,N{=}9728$, \texttt{fp16} & \us{74.8} & \us{77.8} & \us{75.9} & \us{76.8} & \x{1.03} \\
$K{=}2560,N{=}9728$, int4          & \us{24.6} & \us{24.6} & \us{24.6} & \us{27.6} & \x{1.12} \\
$K{=}4096,N{=}12288$, \texttt{fp16}& \us{125.9}& \us{131.1}& \us{131.1}& \us{131.1}& \x{1.04} \\
$K{=}4096,N{=}12288$, int4         & \us{42.0} & \us{42.0} & \us{43.0} & \us{44.0} & \x{1.05} \\
\bottomrule
\end{tabular}
\caption{RTX~4090, PyTorch's native int4 kernel as the stock 4-bit path. A batch
of eight costs \SIrange{5}{12}{\percent} more than a batch of one, against
$4.9$--$5.1\times$ for MLX's tiled GEMM and $4.1$--$4.5\times$ for ours
(\S\ref{sec:marlin:results}).}
\label{tab:nvidia-w4a16}
\end{table}

The consequence for \S\ref{sec:marlin:smallm} is a narrower claim than
``Marlin's premise holds again at $M\geq2$''. NVIDIA's stock 4-bit kernel already
serves $M\leq8$ for almost nothing, so our small-$M$ path closes a gap specific to
MLX's quantised GEMM rather than recovering an instruction-level mechanism. The
speed-up is real and the mechanism is weight reuse; what it demonstrates is a
runtime deficiency, not a vindication of the ported machinery.

\paragraph{Decode attention}
\label{sec:nvidia:attn}

\begin{table}[htbp]
\centering
\small
\begin{tabular}{@{}lrrrrrr@{}}
\toprule
& \multicolumn{3}{c}{Qwen3-4B} & \multicolumn{3}{c}{Qwen3-8B} \\
\cmidrule(lr){2-4}\cmidrule(l){5-7}
\textbf{ctx} & SDPA & FlashInfer & ratio & SDPA & FlashInfer & ratio \\
\midrule
1k  & \us{73.7}  & \us{47.6}  & \x{1.55} & \us{73.7}  & \us{48.1}  & \x{1.53} \\
5k  & \us{77.8}  & \us{53.2}  & \x{1.46} & \us{80.9}  & \us{53.2}  & \x{1.52} \\
10k & \us{100.4} & \us{74.8}  & \x{1.34} & \us{104.4} & \us{76.8}  & \x{1.36} \\
16k & \us{126.0} & \us{101.4} & \x{1.24} & \us{124.9} & \us{102.4} & \x{1.22} \\
\bottomrule
\end{tabular}
\caption{FlashInfer's own decode kernel against PyTorch SDPA on the RTX~4090,
one query per head, KV caches rotated past L2. The gain \emph{falls} with context
where the Metal-port/MLX ratio \emph{rises} on M4~Pro
(Fig.~\ref{fig:kernel-results}); different baselines make this a stack-level rather
than hardware-only comparison. At 1k both kernels are far below the bandwidth
roof --- \gbs{57} and \gbs{88} against \gbs{1008} peak --- so fixed costs and
limited occupancy dominate rather than DRAM bandwidth.}
\label{tab:nvidia-attn}
\end{table}

\paragraph{Speculative decoding}
\label{sec:nvidia:spec}

\begin{table}[htbp]
\centering
\small
\begin{tabular}{@{}lrr@{}}
\toprule
\textbf{Configuration} & \textbf{tok/s} & \textbf{vs.\ autoregressive} \\
\midrule
autoregressive & 150.6 & --- \\
$k=2$ & 117.2 & \x{0.778} \\
$k=4$ & 108.1 & \x{0.717} \\
$k=7$ & 105.9 & \x{0.703} \\
$k=8$ &  86.0 & \x{0.571} \\
\bottomrule
\end{tabular}
\caption{RTX~4090, vLLM, 4-bit Qwen3-8B target with a 4-bit Qwen3-0.6B draft, 32
prompts of 256 generated tokens at \texttt{max\_num\_seqs=1}. At $k=7$ this path
reaches \x{0.703}, versus \x{0.99} with stock speculation on Apple and \x{1.13}
after the small-$M$ Metal kernel accelerates verification
(Table~\ref{tab:marlin-spec}). The nominally matched configuration therefore
changes sign across the tested stacks; differing runtimes and verifier kernels
prevent attributing that change to hardware alone.}
\label{tab:nvidia-spec}
\end{table}

\paragraph{What we did not measure}
\label{sec:nvidia:gaps}

Two comparisons were attempted and are withheld rather than reported.

\textbf{Marlin itself.} vLLM's \texttt{marlin\_gemm} returned an identical
\us{57.3} at every $M$ and at both layer widths --- impossible for a
bandwidth-bound kernel reading \SI{12.45}{\mega\byte} and \SI{3.3}{\mega\byte}
respectively --- which indicates a layout mismatch between the packing helper and
that entry point. Our harness now checks the kernel's output against
$x\,W_{\mathrm{ref}}$ and refuses to report a timing that does not match, so the
NVIDIA side of Table~\ref{tab:nvidia-w4a16} is the stock int4 path only. The
comparison against MLX is stock-to-stock, which is the appropriate pairing.

\textbf{The MoE gather.} At batch~1 our NVIDIA comparison issues one kernel per
(row, expert) pair, so at $M{=}8$ it carries 32 launches
$\approx\us{258}$ of the \us{790} measured --- both the gather and the
grouped-GEMM shape run at \SIrange{200}{260}{\giga\byte\per\second} against
\gbs{900} achievable, i.e.\ both are launch-bound and the comparison between them
is uninformative. A fused-MoE comparison is future work; the
\NotApplicable{} verdict of \S\ref{sec:moe:transfer} rests on the Apple-side
prefill result instead.

\subsubsection{Conclusions}
\label{sec:metal:conclusions}

Portability depends on whether the bottleneck and workload survive the move.
Marlin's weight reuse helps at $M>1$, making it useful for batching and
speculative verification, but not for bandwidth-bound $M=1$. FlashInfer's
split-KV is more relevant on newer Apple GPUs that can repay its extra dispatch
and partials traffic. MoE gather is positive on every tested machine and context
because it removes machinery unused at batch~1. Public Metal's missing
asynchronous copy remains a common gap, and the RTX controls show that CUDA
outcomes do not reliably predict results on Apple Silicon.


\subsection{Quantization}
\label{sec:quantization}

Post-training quantization reduces the model's memory footprint by lowering weight precision after training, trading a small loss in quality for faster inference and lower hardware requirements. We evaluate weight-only post-training quantization of LFM2.5-8B-A1B~\cite{liquidAI20268BA1B, liquidai2025lfm2, smith2026inplace} across three independent toolchains: GGUF/llama.cpp, MLX, and a custom \texttt{llm-compressor}~\cite{llmcompressor2024} pipeline covering RTN, QuaRot~\cite{ashkboos_quarot_2024}+RTN, and QuaRot+GPTQ~\cite{frantar_gptq_2023}, all served on MLX kernels. We benchmark latency and throughput (TTFT, prefill, decode, E2E) for all three toolchains on M2 Pro and M3 Ultra. For cross-platform comparison, we additionally run the GGUF configurations on an NVIDIA RTX~4090. Peak memory and downstream quality (MMLU~\cite{hendryckstest2021, hendrycks2021ethics}, IFBench~\cite{pyatkin2025generalizing}) are reported once per configuration, including the BF16 baselines.

\subsubsection{Latency and throughput}
\label{sec:latency-and-throughput}

Table~\ref{tab:m2pro-long} reports the long-scenario latency and throughput on M2 Pro, and Table~\ref{tab:m3ultra-long} on M3 Ultra. Short-scenario results (512-token prompt / 256-token generation) for both chips are in Appendix~\ref{sec:appendix-quant-perf}. Throughout this section, \emph{family} refers to a group of quantization methods (GGUF, MLX-native, and RTN/QuaRot), while \emph{runtime} refers to the engine that executes them --- llama.cpp for GGUF and MLX for both MLX-native and RTN/QuaRot configurations. Bolding in Tables~\ref{tab:m2pro-long} and~\ref{tab:m3ultra-long} is per runtime.

\begin{table*}[htbp]
\centering
\setlength{\tabcolsep}{5pt}
\renewcommand{\arraystretch}{1.3}
\begin{tabular}{lcccc}
\toprule
Method & TTFT (ms) $\downarrow$ & Prefill (tok/s) $\uparrow$ & Decode (tok/s) $\uparrow$ & E2E (tok/s) $\uparrow$ \\
\midrule
GGUF-Q8\_0    & $914.6$ \std{2.1}   & $1{,}120$ \std{3} & $79.0$ \std{0.8}  & $69.3$ \std{0.6} \\
GGUF-Q6\_K    & $1{,}023.4$ \std{1.8} & $1{,}001$ \std{2} & $93.4$ \std{1.0}  & $78.8$ \std{0.7} \\
GGUF-Q5\_K\_M & $1{,}105.4$ \std{2.0} & $926$ \std{2}    & $88.4$ \std{1.0}  & $74.3$ \std{0.8} \\
GGUF-Q4\_K\_M & $989.7$ \std{1.8}   & $1{,}035$ \std{2}  & $109.9$ \std{1.5} & $90.8$ \std{1.0} \\
GGUF-Q4\_0    & $\mathbf{868.1}$ \std{1.1} & $\mathbf{1{,}180}$ \std{2} & $\mathbf{129.3}$ \std{1.2} & $\mathbf{106.2}$ \std{0.8} \\
\midrule
MLX-W8        & $1{,}901.3$ \std{13.3} & $539$ \std{2} & $84.0$ \std{0.2}  & $63.8$ \std{0.4} \\
MLX-W4        & $1{,}868.1$ \std{6.1}  & $548$ \std{1} & $115.9$ \std{0.5} & $79.9$ \std{3.0} \\
MLX-W4-DWQ    & $1{,}863.0$ \std{3.6}  & $550$ \std{1} & $\mathbf{119.9}$ \std{0.5} & $\mathbf{83.5}$ \std{0.3} \\
MLX-W4-Affine & $1{,}862.1$ \std{5.7}  & $550$ \std{1} & $119.8$ \std{0.6} & $82.7$ \std{1.5} \\
MLX-MXFP4     & $1{,}777.2$ \std{6.4} & $576$ \std{1} & $115.9$ \std{0.5} & $82.7$ \std{0.3} \\
MLX-NVFP4     & $1{,}815.7$ \std{4.6} & $564$ \std{1} & $113.7$ \std{0.3} & $81.1$ \std{0.2} \\
\midrule
RTN-Asym      & $\mathbf{1{,}714}$ \std{4.6} & $\mathbf{597}$ \std{4} & $113.4$ \std{0.7} & $82.3$ \std{0.4} \\
QR+RTN-Asym   & $1{,}730.6$ \std{12.5} & $592$ \std{4} & $110.4$ \std{0.8} & $79.2$ \std{3.1} \\
QR+GPTQ-Asym  & $1{,}733.9$ \std{9.0}  & $591$ \std{3} & $110.3$ \std{0.8} & $80.2$ \std{0.6} \\
QR+GPTQ-Sym   & $1{,}728.6$ \std{6.3}  & $592$ \std{2} & $110.4$ \std{0.7} & $80.1$ \std{1.4} \\
QR+GPTQ-Mixed       & $1{,}728.2$ \std{6.3}  & $593$ \std{2} & $110.3$ \std{0.8} & $80.5$ \std{0.4} \\
\bottomrule
\end{tabular}
\caption{M2 Pro, long scenario (1{,}024-token prompt / 512-token generation): latency and throughput for LFM2.5-8B-A1B. Bold marks the per-column best within each runtime: llama.cpp for the GGUF rows, MLX for both the MLX and RTN/QuaRot (QR) rows ($\downarrow$ lower is better, $\uparrow$ higher is better). The BF16 checkpoint does not fit in M2 Pro memory, so no baseline is shown. Values are the mean $\pm$ standard deviation over repeated runs, with the deviation in grey.}
\label{tab:m2pro-long}
\end{table*}

\begin{table*}[htbp]
\centering
\setlength{\tabcolsep}{5pt}
\renewcommand{\arraystretch}{1.3}
\begin{tabular}{lcccc}
\toprule
Method & TTFT & Prefill & Decode & E2E \\
       & (ms) $\downarrow$ & (tok/s) $\uparrow$ & (tok/s) $\uparrow$ & (tok/s) $\uparrow$ \\
\midrule
BF16 (baseline, GGUF) & $281.7$ \std{0.7} & $3{,}636$ \std{9} & $140.0$ \std{1.4} & $130.2$ \std{1.3} \\
\midrule
GGUF-Q8\_0    & $278.3$ \std{0.7} & $3{,}680$ \std{10} & $192.9$ \std{2.7} & $174.9$ \std{2.2} \\
GGUF-Q6\_K    & $292.7$ \std{0.5} & $3{,}499$ \std{6}  & $198.1$ \std{2.5} & $178.2$ \std{2.1} \\
GGUF-Q5\_K\_M & $308.6$ \std{0.3} & $3{,}318$ \std{4}  & $191.9$ \std{2.5} & $172.3$ \std{2.0} \\
GGUF-Q4\_K\_M & $282.4$ \std{0.7} & $3{,}626$ \std{8}  & $220.7$ \std{3.2} & $197.1$ \std{2.6} \\
GGUF-Q4\_0    & $\mathbf{268.0}$ \std{0.3} & $\mathbf{3{,}820}$ \std{5} & $\mathbf{232.2}$ \std{3.3} & $\mathbf{207.3}$ \std{2.7} \\
\midrule
BF16 (baseline, MLX) & $344.7$ \std{3.8} & $2{,}971$ \std{20} & $141.2$ \std{0.2} & $129.2$ \std{0.2} \\
\midrule
MLX-W8        & $318.0$ \std{2.2} & $3{,}220$ \std{13} & $191.4$ \std{0.3} & $169.7$ \std{2.3} \\
MLX-W4        & $303.0$ \std{3.6} & $3{,}380$ \std{24} & $240.4$ \std{0.5} & $208.4$ \std{4.2} \\
MLX-W4-DWQ    & $300.7$ \std{4.0} & $3{,}406$ \std{27} & $246.2$ \std{1.5} & $214.8$ \std{1.3} \\
MLX-W4-Affine & $299.8$ \std{3.1} & $3{,}416$ \std{21} & $\mathbf{247.0}$ \std{0.7} & $216.1$ \std{0.5} \\
MLX-MXFP4     & $305.8$ \std{3.7} & $3{,}349$ \std{25} & $245.4$ \std{1.1} & $214.3$ \std{0.9} \\
MLX-NVFP4     & $304.0$ \std{3.8} & $3{,}369$ \std{25} & $241.1$ \std{1.1} & $211.2$ \std{0.9} \\
\midrule
RTN-Asym      & $\mathbf{267.5}$ \std{3.9} & $\mathbf{3{,}829}$ \std{55} & $246.8$ \std{0.8} & $\mathbf{217.7}$ \std{3.5} \\
QR+RTN-Asym   & $270.6$ \std{4.7} & $3{,}786$ \std{64} & $238.3$ \std{1.1} & $208.9$ \std{8.1} \\
QR+GPTQ-Asym  & $269.7$ \std{3.1} & $3{,}797$ \std{43} & $238.9$ \std{1.1} & $212.1$ \std{0.9} \\
QR+GPTQ-Sym   & $270.1$ \std{4.0} & $3{,}792$ \std{55} & $236.6$ \std{1.4} & $209.2$ \std{3.8} \\
QR+GPTQ-Mixed & $269.5$ \std{3.4} & $3{,}800$ \std{48} & $236.5$ \std{1.4} & $210.7$ \std{0.9} \\
\bottomrule
\end{tabular}
\caption{M3 Ultra, long scenario (1{,}024-token prompt / 512-token generation): latency and throughput for LFM2.5-8B-A1B. Conventions as in Table~\ref{tab:m2pro-long}; each runtime has its own BF16 baseline.}
\label{tab:m3ultra-long}
\end{table*}

\begin{figure}[htbp]
\centering
\begin{subfigure}[b]{0.49\textwidth}
    \centering
    \includegraphics[width=\textwidth]{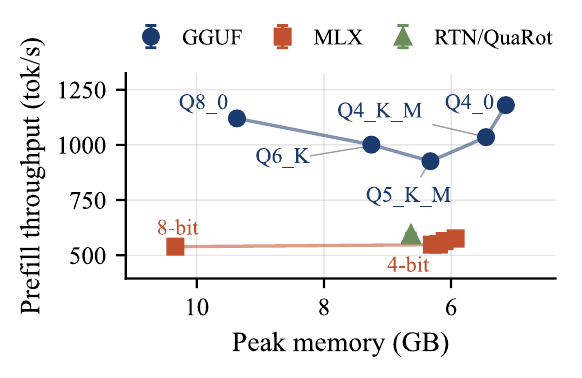}
    \label{fig:m2pro-size-prefill}
\end{subfigure}\hfill
\begin{subfigure}[b]{0.49\textwidth}
    \centering
    \includegraphics[width=\textwidth]{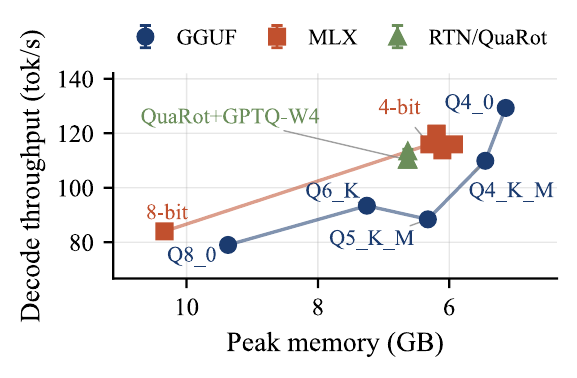}
    \label{fig:m2pro-size-decode}
\end{subfigure}
\caption{M2 Pro: peak memory footprint versus prefill (left) and decode (right) throughput for LFM2.5-8B-A1B, long scenario. The x-axis is reversed, so points further right use less memory and the upper-right corner is best. Full data in Table~\ref{tab:m2pro-long}.}
\label{fig:m2pro-size-speed}
\end{figure}

\paragraph{Decode gains depend on bandwidth headroom.} Figure~\ref{fig:m2pro-size-speed} shows the memory--throughput trade-off on M2 Pro. On both chips, decode throughput rises as bit-width drops, but the extent depends on how memory-bound the chip already is. On M2 Pro, GGUF decode improves from $79.0$ tok/s at Q8\_0 to $129.3$ tok/s at Q4\_0 ($+63.7\%$), and MLX's 4-bit variants reach $113.7$--$119.9$ tok/s. M2 Pro cannot hold the BF16 checkpoint at all, so here quantization is a precondition for running the model rather than an optimization. On M3 Ultra, GGUF-Q4\_0 and MLX-W4-Affine improve decode by $+65.9\%$ and $+74.9\%$ over their BF16 baselines. Relative to 8-bit, however, M3 Ultra gains only $20.4\%$, compared with $63.7\%$ on the more bandwidth-constrained M2 Pro.

\paragraph{Prefill gains less from quantization than decode.}
Prefill benefits far less from lower bit-width than decode (Figure~\ref{fig:m2pro-size-speed}, left, for M2 Pro).
Moving from 8- to 4-bit raises MLX-native prefill by only $2$--$7\%$ on M2 Pro ($539$ to $548$--$576$ tok/s) and $4$--$6\%$ on M3 Ultra ($3{,}220$ to $3{,}349$--$3{,}416$ tok/s), against $20$--$64\%$ for decode, consistent with prefill being compute-bound rather than bandwidth-bound. GGUF prefill, in contrast, depends on the quantization format rather than on the number of bits. All three k-quant formats are slower than Q8\_0 on both chips and in both scenarios: on M2 Pro, Q6\_K, Q5\_K\_M, and Q4\_K\_M reach $1{,}001$, $926$, and $1{,}035$ tok/s against $1{,}120$ tok/s for Q8\_0 (up to $-17\%$), and on M3 Ultra $3{,}499$, $3{,}318$, and $3{,}626$ against $3{,}680$ tok/s (up to $-10\%$). Only the Q4\_0 format exceeds Q8\_0 ($+5\%$ on M2 Pro, $+4\%$ on M3 Ultra). Section~\ref{sec:cross-platform-comparison} shows that this k-quant slowdown is largely absent on the RTX~4090.

\paragraph{The best runtime depends on the chip.} On M2 Pro, GGUF-Q4\_0 is the clear choice. It has the lowest TTFT, the highest prefill, decode, and E2E throughput, and the smallest memory footprint of all configurations. Its prefill is about $2\times$ that of the fastest MLX-runtime configuration, RTN-Asym ($1{,}180$ vs.\ $597$ tok/s), and its decode is $8\%$ higher than that of the best MLX-native variant, MLX-W4-DWQ ($129.3$ vs.\ $119.9$ tok/s).
On M3 Ultra, the two runtimes are close. GGUF-Q4\_0 leads the MLX-native formats on TTFT and prefill ($3{,}820$ vs.\ $3{,}416$ tok/s for MLX-W4-Affine, $+12\%$), while MLX-W4-Affine leads on decode ($247.0$ vs.\ $232.2$ tok/s, $+6\%$). Our RTN-Asym checkpoint, also served on MLX, matches GGUF-Q4\_0 on TTFT and prefill within measurement noise ($267.5$ vs.\ $268.0$~ms and $3{,}829$ vs.\ $3{,}820$ tok/s) and has the highest E2E throughput on the chip ($217.7$ tok/s).

\paragraph{Rankings within each family.} GGUF-Q4\_0 leads the GGUF family on every metric, on both chips and in both scenarios. The MLX-native 4-bit variants are nearly tied. In the long scenario, MLX-MXFP4 leads TTFT and prefill on M2 Pro and MLX-W4-DWQ leads decode and E2E, while on M3 Ultra MLX-W4-Affine leads all four. In the short scenario (Appendix Tables~\ref{tab:appendix-m2pro-short} and~\ref{tab:appendix-m3ultra-short}), the ranking on M2 Pro is largely unchanged: MLX-MXFP4 remains best on TTFT and prefill, while MLX-W4-DWQ and MLX-W4-Affine tie on decode ($121.5$ tok/s) and differ by $0.1$ tok/s on E2E. On M3 Ultra, the best variant differs per metric, and the gaps between them fall within one standard deviation, so their ranking is not robust. Within the RTN/QuaRot family, plain RTN-Asym is fastest on every metric on both chips and in both scenarios. In our weight-only setting, QuaRot's rotations are pure overhead: because RTN and GPTQ differ only during calibration, rotation is the primary runtime variable, and it costs $2.4$--$4.2\%$ of decode and about $1\%$ of long-scenario prefill relative to RTN-Asym. This is expected, since Hadamard transforms yield speedups only when paired with low-precision activation kernels, which this setting does not use.

\subsubsection{Cross-platform comparison}
\label{sec:cross-platform-comparison}

To test which of these trends are specific to Apple Silicon, we run the GGUF configurations on an NVIDIA RTX~4090 and compare them with both Apple chips (Figure~\ref{fig:rtx4090-m2-m3-comparisson}). RTX~4090-only results are in Appendix Figure~\ref{fig:appendix-rtx4090-size-speed}.

\paragraph{The RTX~4090 is fastest at every quantization level.} The ordering RTX~4090 $>$ M3 Ultra $>$ M2 Pro holds at every level for both prefill and decode (Figure~\ref{fig:rtx4090-m2-m3-comparisson}). At Q4\_0, the 4090's decode ($469.1$ tok/s) is $2.0\times$ that of M3 Ultra ($232.2$ tok/s) and $3.6\times$ that of M2 Pro ($129.3$ tok/s). The gap is far wider for prefill: at Q8\_0, the 4090 reaches $13{,}308$ tok/s, $3.6\times$ M3 Ultra and $11.9\times$ M2 Pro.

\paragraph{Relative decode gains differ across devices.}
Normalized to each device's own Q8\_0 (Figure~\ref{fig:rtx4090-m2-m3-comparisson-normalized}), Q4\_0 improves decode by $+63.7\%$ on M2 Pro, $+38.2\%$ on the RTX~4090, and $+20.4\%$ on M3 Ultra. M2 Pro, the most bandwidth-constrained device, benefits the most. M3 Ultra gains the least, even less than the RTX~4090, although both have high memory bandwidth. This suggests that a larger share of its decode time goes to costs that do not depend on bit-width, such as per-token scheduling overheads in llama.cpp's Metal backend.

\begin{figure}[htbp]
\centering
\begin{subfigure}[b]{0.48\textwidth}
\centering
\includegraphics[width=\textwidth]{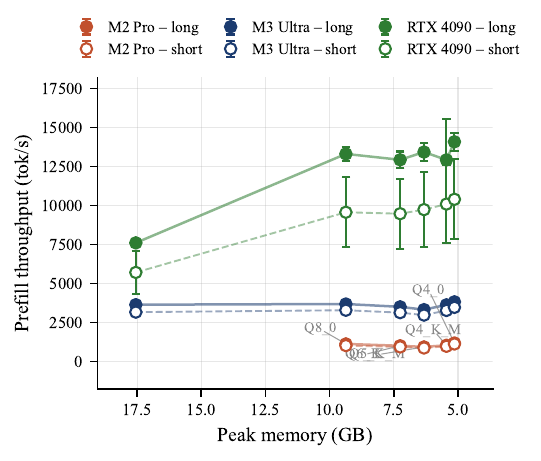}
\caption{Prefill throughput}
\label{fig:rtx4090-m2-m3-comparisson-prefill}
\end{subfigure}
\hfill
\begin{subfigure}[b]{0.48\textwidth}
\centering
\includegraphics[width=\textwidth]{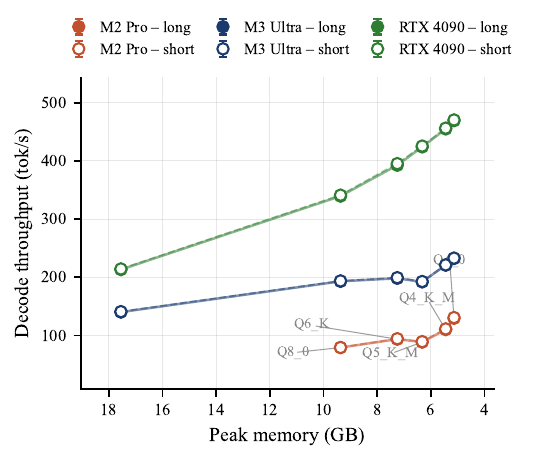}
\caption{Decode throughput}
\label{fig:rtx4090-m2-m3-comparisson-decode}
\end{subfigure}
\caption{GGUF prefill and decode throughput versus peak memory across M2 Pro, M3 Ultra, and RTX 4090. M2 Pro has no BF16 point, as the checkpoint does not fit in its memory. Error bars show standard deviation over 10 repeated runs.}
\label{fig:rtx4090-m2-m3-comparisson}
\end{figure}

\begin{figure}[htbp]
\centering
\begin{subfigure}[b]{0.48\textwidth}
\centering
\includegraphics[width=\textwidth]{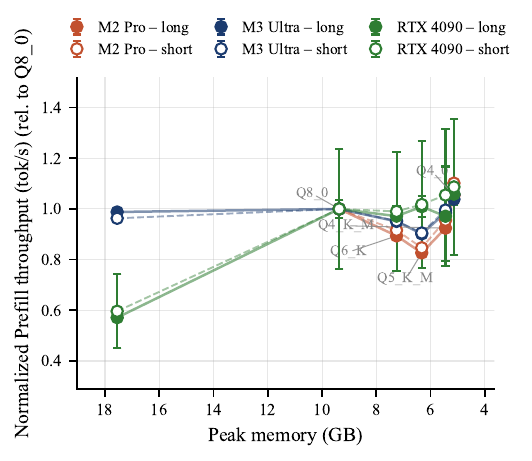}
\caption{Prefill throughput}
\label{fig:rtx4090-m2-m3-comparisson-prefill-normalized}
\end{subfigure}
\hfill
\begin{subfigure}[b]{0.48\textwidth}
\centering
\includegraphics[width=\textwidth]{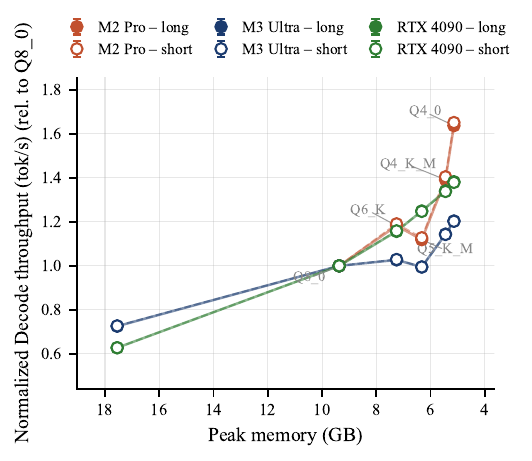}
\caption{Decode throughput}
\label{fig:rtx4090-m2-m3-comparisson-decode-normalized}
\end{subfigure}
\caption{Same data as Figure~\ref{fig:rtx4090-m2-m3-comparisson}, each device's own Q8\_0 throughput normalized to 1.0. Q8\_0 is used as the common reference rather than BF16 because M2 Pro has no BF16 measurement; this normalization isolates the effect of moving from 8-bit to lower bit-widths on each device, independent of each device's absolute throughput.}
\label{fig:rtx4090-m2-m3-comparisson-normalized}
\end{figure}

\paragraph{The k-quant prefill slowdown is specific to Apple Silicon.} On the RTX~4090, Q5\_K\_M prefill is essentially flat relative to Q8\_0 ($+0.9\%$), whereas it falls by $17.3\%$ on M2 Pro and $9.8\%$ on M3 Ultra (\S\ref{sec:latency-and-throughput}). This suggests that the slowdown reflects how llama.cpp's Metal backend executes k-quant formats rather than a property of the formats themselves.

\paragraph{The RTX~4090 shows a prefill jump at 8 bits.} Its prefill throughput rises $1.75\times$ from BF16 to Q8\_0, from $7{,}602$ to $13{,}308$ tok/s (Appendix Figure~\ref{fig:rtx4090-size-prefill}). On M3 Ultra, the same step barely changes prefill ($3{,}636$ to $3{,}680$ tok/s, $1.01\times$). This jump is consistent with Ada Lovelace's INT8 tensor-core paths engaging only once weights are quantized to 8 bits, a mechanism with no direct analog on Apple's GPU architecture.

\subsubsection{Memory footprint}
\label{sec:memory-footprint}
\begin{table*}[htbp]
\centering
\setlength{\tabcolsep}{5pt}
\renewcommand{\arraystretch}{1.3}
\begin{tabular}{lccc}
\toprule
Method & Bits & Model memory (MB) $\downarrow$ & Peak (MB) $\downarrow$ \\
\midrule
BF16 (baseline, GGUF) & 16 & 17{,}463 & 17{,}551 \\
\midrule
GGUF Q8\_0    & 8 & 9{,}280 \pct{$-$46.9\%} & 9{,}368 \pct{$-$46.6\%} \\
GGUF Q6\_K    & 6 & 7{,}167 \pct{$-$59.0\%} & 7{,}254 \pct{$-$58.7\%} \\
GGUF Q5\_K\_M & 5 & 6{,}237 \pct{$-$64.3\%} & 6{,}325 \pct{$-$64.0\%} \\
GGUF Q4\_K\_M & 4 & 5{,}362 \pct{$-$69.3\%} & 5{,}450 \pct{$-$68.9\%} \\
GGUF Q4\_0    & 4 & $\mathbf{5{,}051}$ \pct{$\mathbf{-71.1\%}$} & $\mathbf{5{,}139}$ \pct{$\mathbf{-70.7\%}$} \\
\midrule
BF16 (baseline, MLX) & 16 & 16{,}816 & 18{,}141 \\
\midrule
MLX-W8        & 8 & 9{,}079 \pct{$-$46.0\%} & 10{,}336 \pct{$-$43.0\%} \\
MLX-W4        & 4 & 4{,}815 \pct{$-$71.4\%} & 6{,}302 \pct{$-$65.3\%} \\
MLX-W4-DWQ    & 4 & 4{,}911 \pct{$-$70.8\%} & 6{,}192 \pct{$-$65.9\%} \\
MLX-W4-affine & 4 & 4{,}916 \pct{$-$70.8\%} & 6{,}195 \pct{$-$65.8\%} \\
MLX-MXFP4     & 4 & $4{,}633$ \pct{$-$72.4\%} & $\mathbf{5{,}930}$ \pct{$\mathbf{-67.3\%}$} \\
MLX-NVFP4     & 4 & 4{,}821 \pct{$-$71.3\%} & 6{,}103 \pct{$-$66.4\%} \\
\midrule
RTN-W4-asym         & 4 & $\mathbf{4{,}600}$ \pct{$\mathbf{-72.6\%}$} & 6{,}630 \pct{$-$63.5\%} \\
QuaRot+RTN-W4-asym  & 4 & 4{,}610 \pct{$-$72.6\%} & 6{,}630 \pct{$-$63.5\%} \\
QuaRot+GPTQ-W4-asym & 4 & 4{,}610 \pct{$-$72.6\%} & 6{,}630 \pct{$-$63.5\%} \\
QuaRot+GPTQ-W4-sym  & 4 & 4{,}610 \pct{$-$72.6\%} & 6{,}630 \pct{$-$63.5\%} \\
QuaRot+GPTQ-W4-mixed      & 4 & 4{,}610 \pct{$-$72.6\%} & 6{,}640 \pct{$-$63.4\%} \\
\bottomrule
\end{tabular}
\caption{Memory footprint across all evaluated quantization methods for LFM2.5-8B-A1B. Bold marks the per-column best within each runtime; percentages are relative to that runtime's BF16 baseline. GGUF's KV cache is omitted because it is effectively constant across quantization levels ($12.6$~MB at 512 tokens, $22.0$~MB at 1{,}024 tokens).}
\label{tab:memory}
\end{table*}

Table~\ref{tab:memory} reports model and peak memory for every configuration. Memory footprint decreases roughly in proportion to bit-width across all methods (GGUF-Q4\_0: $-70.7\%$, MLX-MXFP4: $-67.3\%$). Within each runtime's native family, the method with the lowest peak memory also has the smallest weights (GGUF-Q4\_0 and MLX-MXFP4). RTN/QuaRot's peak memory ($6{,}630$--$6{,}640$~MB) is notably higher than that of any MLX 4-bit variant ($5{,}930$--$6{,}302$~MB), despite comparable or smaller model-memory footprints ($4{,}600$--$4{,}610$~MB vs.\ $4{,}633$--$4{,}916$~MB). The gap between model memory and peak memory is roughly $2{,}020$--$2{,}030$~MB for RTN/QuaRot, compared with $1{,}279$--$1{,}487$~MB for MLX's 4-bit variants. This points to a larger transient memory overhead in our RTN/QuaRot pipeline, beyond the quantized weights themselves. Because plain RTN-Asym, which uses neither rotation nor calibration, shows the same gap, the overhead likely arises from how the pipeline loads or converts checkpoints for MLX rather than from QuaRot itself.

\subsubsection{Downstream quality}
\label{sec:downstream-quality}
Table~\ref{tab:quality} reports MMLU and IFBench scores for every configuration. All MMLU scores carry a 95\% confidence interval of $\pm 0.8$ points, and all IFBench scores a 95\% confidence interval of $\pm 5.6$ points, reflecting sampling uncertainty from the fixed evaluation-set size rather than run-to-run measurement variance. These intervals are constant across methods and are therefore omitted from the table.

\begin{table*}[htbp]
\centering
\setlength{\tabcolsep}{5pt}
\renewcommand{\arraystretch}{1.3}
\begin{tabular}{lccc}
\toprule
Method & MMLU $\uparrow$ & IFBench $\uparrow$ & Average $\uparrow$ \\
\midrule
BF16 (baseline, GGUF) & $55.9$ & $55.7$ & $55.8$ \pct{+0.0\%} \\
\midrule
GGUF Q8\_0    & $55.7$ \pct{$-$0.3\%} & $58.3$ \pct{+4.8\%} & $57.0$ \pct{+2.2\%} \\
GGUF Q6\_K    & $\mathbf{56.0}$ \pct{$\mathbf{+0.1\%}$} & $\mathbf{60.7}$ \pct{$\mathbf{+9.0\%}$} & $\mathbf{58.4}$ \pct{$\mathbf{+4.6\%}$} \\
GGUF Q5\_K\_M & $55.6$ \pct{$-$0.5\%} & $55.0$ \pct{$-$1.3\%} & $55.3$ \pct{$-$0.9\%} \\
GGUF Q4\_K\_M & $54.2$ \pct{$-$3.0\%} & $55.0$ \pct{$-$1.3\%} & $54.6$ \pct{$-$2.2\%} \\
GGUF Q4\_0    & $54.0$ \pct{$-$3.4\%} & $53.3$ \pct{$-$4.3\%} & $53.7$ \pct{$-$3.9\%} \\
\midrule
BF16 (baseline, MLX) & $55.68$ & $51.67$ & $53.7$ \pct{+0.0\%} \\
\midrule
MLX-W8        & $\mathbf{55.71}$ \pct{$\mathbf{+0.1\%}$} & $52.67$ \pct{+1.9\%} & $54.2$ \pct{+1.0\%} \\
MLX-W4        & $53.48$ \pct{$-$4.0\%} & $\mathbf{57.33}$ \pct{$\mathbf{+11.0\%}$} & $\mathbf{55.4}$ \pct{$\mathbf{+3.2\%}$} \\
MLX-W4-DWQ    & $54.51$ \pct{$-$2.1\%} & $52.33$ \pct{+1.3\%} & $53.4$ \pct{$-$0.5\%} \\
MLX-W4-affine & $53.20$ \pct{$-$4.5\%} & $53.33$ \pct{+3.2\%} & $53.3$ \pct{$-$0.8\%} \\
MLX-MXFP4     & $52.35$ \pct{$-$6.0\%} & $52.67$ \pct{+1.9\%} & $52.5$ \pct{$-$2.2\%} \\
MLX-NVFP4     & $52.54$ \pct{$-$5.6\%} & $55.67$ \pct{+7.7\%} & $54.1$ \pct{+0.8\%} \\
\midrule
RTN-W4-asym         & $52.96$ \pct{$-$4.9\%} & $49.00$ \pct{$-$5.2\%} & $51.0$ \pct{$-$5.0\%} \\
QuaRot+RTN-W4-asym  & $51.88$ \pct{$-$6.8\%} & $42.00$ \pct{$-$18.7\%} & $46.9$ \pct{$-$12.5\%} \\
QuaRot+GPTQ-W4-asym & $53.09$ \pct{$-$4.7\%} & $41.00$ \pct{$-$20.7\%} & $47.0$ \pct{$-$12.4\%} \\
QuaRot+GPTQ-W4-sym  & $52.94$ \pct{$-$4.9\%} & $39.60$ \pct{$-$23.4\%} & $46.3$ \pct{$-$13.8\%} \\
QuaRot+GPTQ-W4-mixed      & $53.50$ \pct{$-$3.9\%} & $40.30$ \pct{$-$22.0\%} & $46.9$ \pct{$-$12.6\%} \\
\bottomrule
\end{tabular}%
\caption{Downstream quality (MMLU, IFBench) across all evaluated quantization methods for LFM2.5-8B-A1B ($\uparrow$ higher is better). Bold indicates the best value within each runtime. Percentages are relative to the BF16 runtime baseline for each method (MLX for both MLX-native and RTN/QuaRot). Average is the unweighted mean of MMLU and IFBench.}
\label{tab:quality}
\end{table*}

GGUF-Q6\_K has the highest average quality in either runtime ($58.4$), and GGUF-Q8\_0 also scores above its BF16 baseline on average. Their gains, however, fall within the confidence intervals ($+0.1$ MMLU and $+5.0$ IFBench points for Q6\_K), so these configurations match rather than exceed BF16. On the MLX side, MLX-W4 achieves the best average ($+3.2\%$), driven by an IFBench gain of $5.7$ points, at the edge of the confidence interval, that outweighs its MMLU loss of $2.2$ points. Across both runtimes, the GGUF and MLX-native 4-bit variants lose $1.2$--$3.3$ MMLU points relative to their BF16 baseline.
The RTN/QuaRot family behaves differently on the two benchmarks. On MMLU, its scores ($51.9$--$53.5$) are comparable to those of the MLX-native 4-bit variants ($52.4$--$54.5$). On IFBench, every RTN/QuaRot variant scores below every GGUF and MLX-native configuration. Plain RTN-W4-asym loses $2.7$ points against the MLX BF16 baseline, within the confidence interval, whereas the four QuaRot variants lose $9.7$--$12.1$ points. Although IFBench carries a much wider confidence interval than MMLU ($\pm 5.6$ vs.\ $\pm 0.8$ points), the QuaRot drop exceeds it and is consistent across all four variants, which argues against the effect being noise.

\subsection{Conclusions}
\label{sec:quant-conclusions}
As with the kernel results in \S\ref{sec:metal-kernels}, a reported quantization benefit does not automatically carry over across platforms. The same Q8\_0-to-Q4\_0 step speeds up GGUF decode by $1.64\times$ on M2 Pro but only $1.20\times$ on M3 Ultra, and the k-quant prefill slowdown on both Apple chips is largely absent on the RTX~4090. Without low-precision activation kernels, QuaRot's rotations are pure overhead, and in our pipeline they also reduce IFBench quality. The best choice is chip-dependent: GGUF-Q4\_0 dominates on M2 Pro, while on M3 Ultra it leads on prefill and MLX's affine/DWQ variants lead on decode.

\section{Discussion and Conclusion}
Across four optimization families, no single verdict holds uniformly. Each technique's fate on Apple Silicon depends on a specific, identifiable mismatch between its assumptions and our batch-1, unified-memory regime. 

\textbf{KV-cache pruning} helps only when the selection cost is low relative to the attention it removes. Profit grows with output length, context, and architecture, and reverses on NVIDIA GPUs. \textbf{Speculative decoding} produced no standardized E2E gain in the principal long-prompt chat benchmark. Draft–verify overhead and prefill penalties offset its decode-side benefits, although separate configurations showed positive results. \textbf{Metal kernel} ports succeed only where the workload matches their design. Marlin's weight reuse pays off only once more than one token decodes at a time, useful for batching and draft verification but not single-token chat; FlashInfer's split-KV attention earns back its overhead only on newer Apple GPUs; and the MoE gather helps everywhere, since it removes scheduling work batch-one decoding never needed. \textbf{Quantization} gains reflect how memory-bound the chip already is, rather than bit-width itself. Neither GGUF nor MLX is better in all scenarios. Generally, GGUF is faster on prefill, but MLX has better decode speed, so the best choice depends on which phase matters more for the workload.

The unifying result is that a reported speedup is not a portable property of a technique alone, but a joint function of technique, runtime, hardware contract, and workload regime. Practitioners porting CUDA-validated techniques to Apple Silicon should identify which bottleneck a mechanism actually removes before expecting a number.

\textbf{Limitations.} Our KV-pruning lacks task-quality evaluation, and the megakernel study covers a single model on a single chip. Extending this analysis to multi-request serving, where batching gives several currently idle mechanisms something to act on, is left to future work.

\bibliographystyle{unsrtnat}
\bibliography{references_zotero,references}


\clearpage

\appendix

\section{Additional Speculative-Decoding Results}
\label{app:speculative-details}

This appendix records the implementation settings and supplementary 8B sensitivity experiment. The method review, acceptance/cost analysis, limiting factors, and configuration-specific positive cases appear in Section~\ref{sec:speculative-decoding}.

\subsection{Implementation and benchmark configuration}
\label{app:speculative-implementation}

The principal comparison uses MLX with batch size one and concurrency one on M2 Pro and M3 Ultra. Targets are \texttt{mlx-community/Qwen3-4B-4bit} and BF16 \texttt{Qwen/Qwen3-4B}. The 100-prompt conversational dataset contains 46 tool-calling and 54 regular-chat samples, with a mean prefill length of \(1{,}776\) tokens. Baseline completions average \(61.9\) tokens with the 4-bit target and \(54.2\)--\(55.0\) tokens with BF16. Generation is greedy (\(T=0\)), with a 512-token limit, one warmup, and three measured repetitions per prompt and method. Method-specific settings are held fixed across the two devices and target precisions unless stated below.

\paragraph{Classic.}
The auxiliary autoregressive drafter is \texttt{mlx-community/Qwen3-0.6B-4bit} for both target precisions. The draft width is four tokens, the fastest setting in our tested configurations.

\paragraph{N-gram and Cascade-n-gram.}
N-gram selects the longest matching stored suffix from a precomputed lookup table, with match lengths from three to eight tokens, and proposes two tokens per round. Cascade-n-gram uses the same table to speculate for the 0.6B-4bit model; its verified predictions then serve as drafts for the target. The target-level draft width remains two tokens. This implementation evaluates a vertical N-gram cascade, rather than the full vertical and horizontal Max-Gram cascade of~\citet{chen2024cascade}.

\paragraph{EAGLE-3.}
The community head \texttt{AngelSlim/Qwen3-4B\_eagle3} is converted to MLX and quantized to four bits, including for the BF16-target runs. It takes hidden states from target layers \(2\), \(18\), and \(33\). One token is drafted per step, the fastest tested setting; this local configuration differs from the wider dynamic-tree settings used to obtain the published speedups.

\paragraph{DFlash and DDTree.}
Both use the BF16 checkpoint \texttt{z-lab/Qwen3-4B-DFlash-b16}~\cite{zlab2026qwen3dflash}. DFlash's 16-position block contains the target-produced root and up to 15 draft candidates. DDTree builds a tree with a budget of 24 non-root nodes from the same block drafter, the best fixed budget in the implementation's local sweep. Both use greedy verification.

\paragraph{Baseline and acceptance conventions.}
Each adaptation is compared with its own autoregressive baseline. Standardized E2E throughput uses the corresponding baseline token workload, so speculative output-length changes do not redefine the workload used for this metric. Percentage changes are computed relative to these paired baselines; absolute rates across different implementations are not a controlled cross-method comparison. In Figure~\ref{fig:spec_decoding}(c), AL counts accepted draft tokens and excludes the target-produced root for DFlash and DDTree. DT denotes the configured budget: four, two, two, and one draft tokens for Classic, N-gram, Cascade-n-gram, and EAGLE-3; a 16-position block including the root for DFlash; and 24 non-root tree nodes for DDTree. These differing budget definitions must be respected when interpreting AL. The committed-token count \(\tau\) in Equation~\ref{eq:speculative-cost} includes target-produced progress as applicable and is not the same metric as AL.

\subsection{Additional results for Qwen3-8B-4bit}
\label{app:speculative-8b}

\begin{table}[htbp]
    \centering
    \small
    \setlength{\tabcolsep}{3pt}
    \caption{Standardized end-to-end (E2E), prefill, and decode throughput changes relative to each implementation's autoregressive baseline for Qwen3-8B-4bit on M2 Pro and M3 Ultra.}
    \label{tab:speculative-apple-8b-4bit}
    \begin{tabular}{@{}lrrrrrr@{}}
        \toprule
        & \multicolumn{3}{c}{M2 Pro} & \multicolumn{3}{c}{M3 Ultra} \\
        \cmidrule(lr){2-4} \cmidrule(l){5-7}
        Method & E2E & Prefill & Decode & E2E & Prefill & Decode \\
        \midrule
        Classic
            & \(-14.4\%\) & \(-9.1\%\) & \(-38.6\%\)
            & \(-16.9\%\) & \(-15.1\%\) & \(-22.2\%\) \\
        N-gram
            & \(-4.6\%\) & \(-0.4\%\) & \(-25.6\%\)
            & \(-1.4\%\) & \(-0.4\%\) & \(-4.4\%\) \\
        Cascade-n-gram
            & \(-12.6\%\) & \(-9.8\%\) & \(-27.2\%\)
            & \(-10.2\%\) & \(-10.4\%\) & \(-9.6\%\) \\
        EAGLE-3
            & \(-14.6\%\) & \(-10.5\%\) & \(-34.3\%\)
            & \(-12.5\%\) & \(-9.7\%\) & \(-20.4\%\) \\
        DFlash
            & \(-28.4\%\) & \(-10.0\%\) & \(-69.4\%\)
            & \(-29.5\%\) & \(-8.7\%\) & \(-59.5\%\) \\
        DDTree
            & \(-23.8\%\) & \(-10.0\%\) & \(-62.0\%\)
            & \(-26.6\%\) & \(-8.7\%\) & \(-55.1\%\) \\
        \bottomrule
    \end{tabular}
\end{table}

To test whether a more expensive target step makes speculation more favorable, we repeat the M2 Pro and M3 Ultra comparison with \texttt{Qwen/Qwen3-8B-MLX-4bit}. The learned checkpoints are replaced by \texttt{AngelSlim/Qwen3-8B\_eagle3} and \texttt{z-lab/Qwen3-8B-DFlash-b16}; the Classic drafter and other method parameters are unchanged. The workload, decoding protocol, and paired-baseline normalization follow the principal comparison.

Increasing the target from 4B to 8B improves relative E2E performance for most configurations, but no method achieves an E2E gain. N-gram comes closest on M3 Ultra (\(-1.4\%\)). Decode changes are method-dependent: EAGLE-3 becomes slower relative to its baseline on both devices, whereas DFlash and DDTree modestly reduce their penalties. The larger target therefore provides additional amortization headroom without eliminating the drafting, verification, and prefill costs identified in the main text.

\clearpage

\section{Appendix Metal}

\subsubsection{Marlin: 6-bit \texttt{lm\_head}}
\label{app:marlin:lmhead}

A 6-bit matvec for the tied \texttt{lm\_head}, using a $4{+}2$ bit-plane layout,
measured parity with stock MLX (\us{835} versus \us{844}). MLX sustained
\gbs{252}--\gbs{264} on that shape at 4, 6 and 8 bits alike, showing that unpack
cost was not the limiting resource: the shape is rate-capped, so a cheaper
unpack has nothing to recover (\S\ref{sec:marlin:results}).

\subsubsection{FlashInfer: split-KV routing calibration}
\label{app:fi:routing}

FlashInfer-style implementations gate split-KV on a context threshold
\cite{ye_flashinfer_2025}. The reference port used $N\geq1024$ on every GPU,
forcing a two-pass path where stock MLX remained single-pass and under-splitting
by \x{2} at 5--7k. This cost $1$--$3\%$ on M2. Our router
(Listing~\ref{lst:fi-routing}) therefore uses a measured per-architecture table;
an unmeasured machine inherits stock behaviour rather than a heuristic tuned on
different hardware.

\begin{metalcode}
// cider_mlx/sdpa.py -- the routing decision, resolved once at import
bool use_2pass(int N, int gqa) {
    if (!use_mlx_routing) return N >= 1024;        // the reference heuristic
    if (device_class == 's' || device_class == 'd') return N >= 1024;
    return gqa > 1 && N >= 4096;                   // MLX's own gate
}
// Architectures measured to prefer MLX's routing over the reference:
//   applegpu_g14g (M2) -- the flat threshold forced split-KV where stock
//   stays 1-pass, and under-split 2x at 5-7k: -1..-3%.
\end{metalcode}
\captionof{listing}{Split-KV is enabled by measurement, not by default. An
architecture that nobody has benchmarked inherits stock behaviour rather than a
heuristic tuned elsewhere.}
\label{lst:fi-routing}

\subsubsection{Megakernel: cooperation, occupancy and geometry}
\label{app:mega}

The probes below complete the exploratory case study of \S\ref{sec:mega}. As
there, they cover one LFM2.5-8B-A1B MoE block on one M4~Pro under MLX.

\paragraph{P2: probing each cooperation guarantee}
\label{app:mega:p2}

Each clause CUDA guarantees in writing, probed on an M4~Pro:

\begin{center}\small
\begin{tabular}{@{}lll@{}}
\toprule
\textbf{CUDA guarantee} & \textbf{Public Metal} & \textbf{Measured} \\
\midrule
Cooperative launch    & none              & 1024 toy TGs resident; \textbf{2048 silently fail} \\
\ccode{grid.sync()}   & hand-rolled       & works: \us{0.38} at 256-thread TGs \\
Fenced plain stores   & none              & \SI{96}{\percent} stale across TGs \\
Scoped atomics        & relaxed only      & coherent: 0 stale in \num{41.3e6} reads \\
\ccode{cp.async}      & none              & no prefetch across a barrier \\
\bottomrule
\end{tabular}
\end{center}

\noindent
Rows two and four are why a barrier can be built at all
(Listing~\ref{lst:mega-barrier}); rows one and three are why it cannot be trusted.
In this probe, ordinary device writes were not reliably visible across
threadgroups and \mcode{volatile} did not fix them, so each tested cross-phase
handoff had to round-trip through relaxed atomics.

\begin{metalcode}
// The only grid barrier Metal permits: monotonic counter, relaxed atomics,
// bounded spin. Correct only if EVERY threadgroup is already resident.
threadgroup_barrier(mem_flags::mem_threadgroup);        // 1. sync within my group
if (thread_index_in_threadgroup == 0) {                 // 2. one vote per group
    atomic_fetch_add_explicit(counter, 1u, memory_order_relaxed);
    while (atomic_load_explicit(counter, memory_order_relaxed) < NUM_TGS) {
        /* 3. spin -- deadlocks if any threadgroup has not been scheduled */
    }
}
threadgroup_barrier(mem_flags::mem_threadgroup);        // 4. release my group
\end{metalcode}
\captionof{listing}{Step 3 is the whole problem. If the grid exceeds what the GPU
holds, the resident threadgroups spin forever waiting for threadgroups that
cannot start until a resident one finishes. There is no API to determine that in
advance.}
\label{lst:mega-barrier}

\paragraph{P3: the same kernel, one register-footprint apart}
\label{app:mega:p3}

\begin{center}\small
\begin{tabular}{@{}lll@{}}
\toprule
\textbf{Variant} & \textbf{Difference} & \textbf{Result} \\
\midrule
v1 & lighter phase B                & barrier converged, \us{183} \\
v2 & register-heavier phase B, otherwise identical & \textbf{non-convergent}, silent \SI{30}{\milli\second} spin timeouts \\
\bottomrule
\end{tabular}
\end{center}

\noindent
A toy barrier held 1024 threadgroups; the real megakernel, carrying
quantised-matmul register pressure and threadgroup-memory tiles, could not hold
64. One variable changed between v1 and v2: the per-thread register footprint,
which the compiler chooses and never reports. There is no occupancy query with
which to predict the outcome, no cooperative launch that would fail loudly
instead, and no diagnostic distinguishing v2's hang from a deadlocked barrier.

\paragraph{Geometry: fused versus shape-optimal grids}
\label{app:mega:geometry}

With the barrier disabled, measuring pure compute:

\begin{center}\small
\begin{tabular}{@{}lrr@{}}
\toprule
\textbf{Configuration} & \textbf{Time} & \textbf{Effective rate} \\
\midrule
Stock 11-op chain (each op its own grid) & \us{96}  & \gbs{225} over \SI{22}{\mega\byte} \\
One-dispatch megakernel, compute only    & \us{174} & \gbs{124} \\
\bottomrule
\end{tabular}
\end{center}

\noindent
Phase B's row partition wants 64 threadgroups; at 64 threadgroups phase A streams
\SI{14.7}{\mega\byte} of gate and up weights at a third of the rate the same code
achieves in its own dispatch. Shape-optimal grids are therefore worth \us{78} on
this block, against the \us{15} that removing every boundary recovers: the two
effects differ by \x{5} and point in opposite directions.

\paragraph{Would an interpreter design escape this?}
\label{app:mega:interpreter}

Mirage-style designs \cite{wu_mirage_2024} replace fixed phases with an
instruction interpreter, which might seem to sidestep the geometry constraint.
For this workload it would retain both observed pressures. An interpreter
contains the code for every instruction type, so its static register allocation
reflects their combined worst case; and without a public
\ccode{cp.async}-class mechanism it cannot prefetch the next instruction's
weights behind the current one's compute.

\clearpage

\section{Full Quantization Benchmark Results}
\label{sec:appendix-quant-perf}
This section reports the measurements underlying Section~\ref{sec:quantization} that are not shown in the main text: latency and throughput for the short scenario (512-token prompt / 256-token generation) on M2 Pro and M3 Ultra (Tables~\ref{tab:appendix-m2pro-short} and~\ref{tab:appendix-m3ultra-short}), and additional GGUF results on the RTX~4090 (\S\ref{sec:quant-cross-platform}). All results are for LFM2.5-8B-A1B. Family, runtime, and bolding conventions follow Section~\ref{sec:latency-and-throughput}.

\subsection{Latency and throughput}

\begin{table*}[htbp]
\caption{M2 Pro, short scenario (512-token prompt / 256-token generation): latency and throughput for LFM2.5-8B-A1B. Conventions as in Table~\ref{tab:m2pro-long}.}
\label{tab:appendix-m2pro-short}
\centering
\setlength{\tabcolsep}{5pt}
\renewcommand{\arraystretch}{1.3}
\begin{tabular}{lcccc}
\toprule
Method & TTFT & Prefill & Decode & E2E \\
       & (ms) $\downarrow$ & (tok/s) $\uparrow$ & (tok/s) $\uparrow$ & (tok/s) $\uparrow$ \\
\midrule
GGUF-Q8\_0    & $505.7$ \std{1.2} & $1{,}012$ \std{2} & $79.1$ \std{0.7}  & $68.6$ \std{0.6} \\
GGUF-Q6\_K    & $552.5$ \std{2.1} & $927$ \std{3}    & $94.2$ \std{1.1}  & $78.5$ \std{0.8} \\
GGUF-Q5\_K\_M & $597.1$ \std{1.0} & $857$ \std{1}    & $89.2$ \std{1.0}  & $74.0$ \std{0.7} \\
GGUF-Q4\_K\_M & $529.3$ \std{1.6} & $967$ \std{3}    & $111.1$ \std{0.9} & $90.6$ \std{0.7} \\
GGUF-Q4\_0    & $\mathbf{459.0}$ \std{1.7} & $\mathbf{1{,}115}$ \std{4} & $\mathbf{130.6}$ \std{1.2} & $\mathbf{106.1}$ \std{0.8} \\
\midrule
MLX-W8        & $1{,}021.1$ \std{5.7} & $501$ \std{2} & $85.2$ \std{0.3} & $63.7$ \std{0.2} \\
MLX-W4        & $1{,}002.2$ \std{5.1} & $511$ \std{2} & $117.7$ \std{0.5} & $80.7$ \std{0.2} \\
MLX-W4-DWQ    & $997.4$ \std{4.3} & $513$ \std{1} & $\mathbf{121.5}$ \std{0.5} & $82.6$ \std{0.2} \\
MLX-W4-Affine & $993.1$ \std{3.6} & $516$ \std{1} & $\mathbf{121.5}$ \std{0.6} & $\mathbf{82.7}$ \std{0.2} \\
MLX-MXFP4     & $\mathbf{951.9}$ \std{6.1} & $\mathbf{538}$ \std{2} & $117.8$ \std{0.6} & $82.0$ \std{0.2} \\
MLX-NVFP4     & $973.9$ \std{4.5} & $526$ \std{2} & $115.4$ \std{0.4} & $80.3$ \std{0.2} \\
\midrule
RTN-Asym      & $967.8$ \std{9.2} & $529$ \std{5} & $114.9$ \std{1.1} & $80.3$ \std{0.6} \\
QR+RTN-Asym   & $977.8$ \std{9.8}  & $524$ \std{5} & $111.9$ \std{1.0} & $78.6$ \std{0.4} \\
QR+GPTQ-Asym  & $974.0$ \std{5.1}  & $526$ \std{3} & $112.1$ \std{0.7} & $78.8$ \std{0.3} \\
QR+GPTQ-Sym   & $974.1$ \std{7.9}  & $526$ \std{4} & $112.1$ \std{0.9} & $78.8$ \std{0.5} \\
QR+GPTQ-Mixed       & $977.2$ \std{10.0} & $524$ \std{5} & $111.7$ \std{0.9} & $78.5$ \std{0.4} \\
\bottomrule
\end{tabular}
\end{table*}

\begin{table*}[htbp]
\caption{M3 Ultra, short scenario (512-token prompt / 256-token generation): latency and throughput for LFM2.5-8B-A1B, quantized with GGUF, MLX, and \texttt{llm-compressor} (RTN/QuaRot). Bolding and other conventions follow Table~\ref{tab:m2pro-long}; each runtime has its own BF16 baseline.}
\label{tab:appendix-m3ultra-short}
\centering
\setlength{\tabcolsep}{5pt}
\renewcommand{\arraystretch}{1.3}
\begin{tabular}{lcccc}
\toprule
Method & TTFT & Prefill & Decode & E2E \\
       & (ms) $\downarrow$ & (tok/s) $\uparrow$ & (tok/s) $\uparrow$ & (tok/s) $\uparrow$ \\
\midrule
BF16 (baseline, GGUF) & $162.1$ \std{0.5} & $3{,}159$ \std{10} & $140.8$ \std{1.4} & $129.7$ \std{1.2} \\
\midrule
GGUF-Q8\_0    & $156.1$ \std{0.4} & $3{,}280$ \std{8}  & $193.6$ \std{2.7} & $173.7$ \std{2.2} \\
GGUF-Q6\_K    & $163.8$ \std{0.6} & $3{,}126$ \std{11} & $199.2$ \std{2.9} & $177.2$ \std{2.4} \\
GGUF-Q5\_K\_M & $172.4$ \std{0.6} & $2{,}969$ \std{11} & $192.6$ \std{2.6} & $171.0$ \std{2.1} \\
GGUF-Q4\_K\_M & $156.8$ \std{0.3} & $3{,}266$ \std{7}  & $221.2$ \std{3.2} & $195.4$ \std{2.5} \\
GGUF-Q4\_0    & $\mathbf{147.8}$ \std{0.5} & $\mathbf{3{,}465}$ \std{13} & $\mathbf{232.8}$ \std{3.4} & $\mathbf{205.8}$ \std{2.7} \\
\midrule
BF16 (baseline, MLX) & $215.5$ \std{3.2} & $2{,}377$ \std{22} & $142.0$ \std{0.2} & $127.2$ \std{0.2} \\
\midrule
MLX-W8        & $196.7$ \std{4.6} & $2{,}605$ \std{37} & $191.9$ \std{0.3} & $167.8$ \std{0.3} \\
MLX-W4        & $187.9$ \std{5.7} & $2{,}727$ \std{49} & $238.8$ \std{2.7} & $203.7$ \std{2.0} \\
MLX-W4-DWQ    & $183.9$ \std{3.3} & $2{,}785$ \std{30} & $247.0$ \std{2.6} & $210.3$ \std{2.0} \\
MLX-W4-Affine & $184.0$ \std{3.9} & $2{,}783$ \std{36} & $248.0$ \std{0.6} & $211.1$ \std{0.4} \\
MLX-MXFP4     & $186.2$ \std{3.3} & $2{,}751$ \std{30} & $248.6$ \std{2.6} & $210.8$ \std{2.0} \\
MLX-NVFP4     & $186.4$ \std{4.7} & $2{,}748$ \std{42} & $243.4$ \std{2.8} & $207.3$ \std{2.2} \\
\midrule
RTN-Asym      & $\mathbf{173.3}$ \std{4.0} & $\mathbf{2{,}957}$ \std{66} & $\mathbf{249.0}$ \std{1.0} & $\mathbf{213.8}$ \std{0.4} \\
QR+RTN-Asym   & $177.1$ \std{4.8} & $2{,}893$ \std{78} & $241.0$ \std{1.2} & $207.2$ \std{0.8} \\
QR+GPTQ-Asym  & $177.6$ \std{6.1} & $2{,}885$ \std{96} & $240.7$ \std{1.7} & $206.9$ \std{1.2} \\
QR+GPTQ-Sym   & $178.3$ \std{6.6} & $2{,}875$ \std{105} & $241.0$ \std{1.4} & $207.0$ \std{0.9} \\
QR+GPTQ-Mixed       & $179.1$ \std{6.7} & $2{,}862$ \std{104} & $239.9$ \std{1.3} & $206.1$ \std{1.1} \\
\bottomrule
\end{tabular}
\end{table*}

\subsection{Cross-platform comparison}
\label{sec:quant-cross-platform}

\begin{figure}[htbp]
\centering
\begin{subfigure}[b]{0.48\textwidth}
\centering
\includegraphics[width=\textwidth]{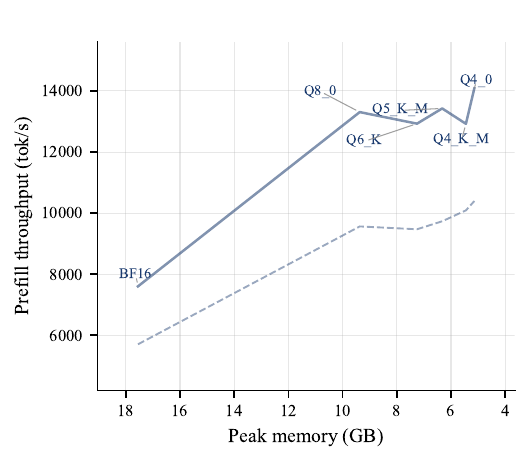}
\caption{Prefill throughput}
\label{fig:rtx4090-size-prefill}
\end{subfigure}
\hfill
\begin{subfigure}[b]{0.48\textwidth}
\centering
\includegraphics[width=\textwidth]{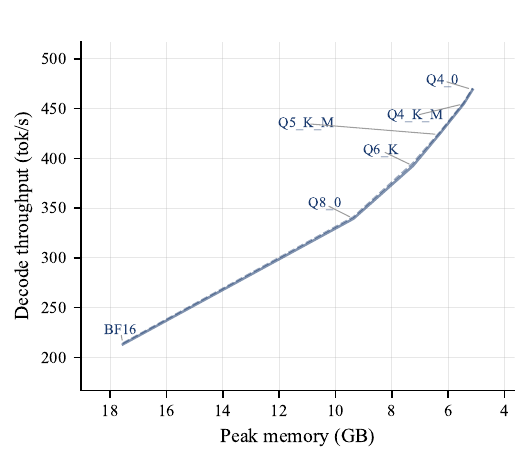}
\caption{Decode throughput}
\label{fig:rtx4090-size-decode}
\end{subfigure}
\caption{RTX~4090: peak memory footprint versus throughput for GGUF quantization of LFM2.5-8B-A1B, same conventions as Figure~\ref{fig:m2pro-size-speed}.}
\label{fig:appendix-rtx4090-size-speed}
\end{figure}

Figure~\ref{fig:appendix-rtx4090-size-speed} shows the RTX~4090 peak memory footprint versus throughput for GGUF quantization of LFM2.5-8B-A1B.

\clearpage

\end{document}